\documentclass{article}
\usepackage[margin=1.2in]{geometry}
\usepackage{graphicx}
\usepackage{inputenc}
\usepackage{multirow}
\usepackage{longtable}
\usepackage{lscape}
\usepackage{multirow}
\usepackage{caption}
\usepackage{float}
\usepackage{newunicodechar}
\newunicodechar{≤}{\ensuremath{\leq}}
\newunicodechar{≥}{\ensuremath{\geq}}
\usepackage{authblk}
\providecommand{\keywords}[1]
{
  \small	
  \textbf{\textit{Keywords---}} #1
}

\usepackage{tcolorbox}
\usepackage[square,sort,comma,numbers]{natbib}

\title {  \begin{tcolorbox} \centering Cite this paper as:\\
H. Hamdoun, A. Sagheer, and H. Youness. ‘Energy Time Series Forecasting-analytical and Empirical Assessment of Conventional and Machine Learning Models’, Journal of Intelligent \& Fuzzy Systems, 40(6), 12477-
12502, 2021 \end{tcolorbox}
{Energy time series forecasting-Analytical and empirical assessment of conventional and machine learning models}}
\author[1,2]{Hala Hamdoun}
\author[1,3]{Alaa Sagheer \thanks{Corresponding author: asagheer@aswu.edu.eg}}
\author[2]{Hassan Youness}

\affil[1]{Center for Artificial Intelligence and RObotics (CAIRO), Faculty of Science, Aswan University, Egypt}
\affil[2]{Department of Computers and Systems Engineering, Faculty of Engineering, Minia University, Egypt}
\affil[2]{College of Computer Science and Information Technology, King Faisal University, Saudi Arabia}
\date{}

\begin{document}

\maketitle

\begin{abstract}
Machine learning methods have been adopted in the literature as contenders to conventional methods to solve the energy time series forecasting (TSF) problems. Recently, deep learning methods have been emerged in the artificial intelligence field attaining astonishing performance in a wide range of applications. Yet, the evidence about their performance in to solve the energy TSF problems, in terms of accuracy and computational requirements, is scanty. Most of the review articles that handle the energy TSF problem are systematic reviews, however, a qualitative and quantitative study for the energy TSF problem is not yet available in the literature. The purpose of this paper is twofold, first it provides a comprehensive analytical assessment for conventional, machine learning, and deep learning methods that can be utilized to solve various energy TSF problems. Second, the paper carries out an empirical assessment for many selected methods through three real-world datasets. These datasets related to electrical energy consumption problem, natural gas problem, and  electric  power consumption of an individual household problem. The first two problems are univariate TSF and the third problem is a multivariate TSF. Compared to both conventional and machine learning contenders, the deep learning methods attain a significant improvement in terms of accuracy and forecasting horizons examined. In the meantime, their computational requirements are notably greater than other contenders. Eventually, the paper identifies a number of challenges, potential research directions, and recommendations to the research community may serve as a basis for further research in the energy forecasting domain.

\end{abstract}

\keywords{Energy time series forecasting,  Conventional forecasting methods, Machine learning, Deep learning, Energy management systems}

\begin{table*}
    \centering

    \begin{tabular}{|llll|}
    \hline
    Abbreviations&&&
\\
TSF&  Time Series Forecasting&SES & Simple Exponential Smoothing\\
ARIMA & Autoregression Integrated Moving Average&MES &  Multivariate Exponential Smoothing\\
ARIMA & Autoregression Moving Average&MLP&  Multilayer Perceptron\\
DCA & Decline Curve Analysis&BP&  backpropagation\\

VAR & Vector Autoregressive&RNN&  Recurrent Neural Networks\\

BVAR&  Bayesian VAR &DRNN&  Deep RNN\\

SVM&  Support Vector Machine &LSTM & Long Short-Term Memory\\

SVR&  Support Vector Regression&DSLTM & Deep LSTM\\

LSSVR & Least Squares SVR&ESN & Echo State Network\\

kNN & k-Nearest Neighbour&MAE & Mean Absolute Error\\

ANN&  Artificial Neural  Networks&MSE&  Mean Square Error\\

UTS & univariate Time Series&RMSE & Root MSE\\

MTS&  Multivariate Time Series&RMSPE & Root Mean Square Percentage Error\\

ES &  Exponential Smoothing&MAPE&  Mean Absolute Percentage Error\\

\hline
    \end{tabular}
\end{table*}
\section{Introduction}
\subsection {Problem overview}
The advent of sensors and measurement technologies has resulted in an exponential growth of time series data that recorded from multiple sources over time \cite{Wei}. This kind of data are rich with dynamical information and has characteristics of temporal dependencies and high dimensionalities. These properties have attracted the practitioners to utilize the time series data to achieve the task of forecasting in a wide spectrum of applications \cite {Douglas}. Actually, forecasting is one of the oldest problems in the human history since the ancient Egyptians who established various mechanisms to measure the Nile river flow to predict flood \cite{Carlos}. However, the emphasis of this paper is on the forecasting using time series data, which will be called henceforth as time series forecasting (TSF) \cite{Srihari}.

TSF is the rational prediction of future trends based on the past and current observations. This is based on the concept that the past observations include intrinsic patterns, which contain related information to the future representation of the problem at hand. This property enables the time series data to capture the causalities of the underlying processes. Therefore, the TSF problem is included in a wide spectrum of applications, including energy.

Energy TSF is one of the most exciting and potentially ground-breaking research field. In the past decades with the dominant usage of traditional energy sources, there were no need for forecasting techniques where the energy demand could be ideally matched with the energy supply. However, in recent decades, the world wide energy markets have drastically grown, particularly with the increasing exploitation of renewable energy sources as well as the intraday real-time trading \cite{Dannecker}. As such, these challenges, and many more, require more accurate and fine-grained forecasting techniques able to predict the energy demands for days and months ahead \cite{Provo}.

Certainly, accurate forecasting of energy supply and demand are an essential requirement for adjusting energy production and consumption and thus, for the stability of the energy markets. However, obtaining reasonably accurate predictions from the energy time series data is quite difficult due to many inevitable limitations. One of the limitations is the fact that the time series notoriously violates the independence and identical distribution property of spatial statistics \cite{Adhikari}. Another limitation is the high volatility and uncertainty of some energy applications, such as household energy load \cite {Shi}. There are other limitations including strong linear dependence between observations, lack of stationarity, and curse of dimensionality \cite{Dannecker}.

Due to these inherent limitations, a limited number of efficient forecasting models have been presented so far, despite sincere research efforts \cite{Parmezan}. For several years, the conventional forecasting methodologies have been widely used to solve the energy TSF \cite {Dannecker}. The conventional methodologies are based on three stochastic parametric methods; namely, autoregression (AR), moving average (MA), and Arps equation. AR and MA were combined together later on to establish a new method called autoregression moving average (ARMA) \cite{Douglas}. ARMA, and its variant autoregression integrated moving average (ARIMA), still represent the widespread techniques used widely to achieve diverse forecasting applications. The Arps equation is the basic of a known forecasting method known as the decline curve analysis (DCA), which has been widely used in the industrial domains \cite{paryani}. Vector autoregressive (VAR) method is also used widely to achieve forecasting using multiple time series \cite{Oliv}. 

In the last decade, with the outstanding progress in AI, research on energy forecasting based on machine learning has been booming. Several machine learning algorithms have drawn a wide attention and presented as serious contenders to conventional methods in the forecasting community due to their favorable performance \cite{mar}. Examples of these approaches are the support vector machine (SVM) \cite{SVM}, the k-nearest neighbour (kNN) \cite{sun}, and the artificial neural networks (ANNs) \cite{Zhang-Book}, and their variations. the are used widely in modeling the energy TSF problems. The inherently nonlinear structure of traditional ANNs, with shallow architectures, is particularly versatile for capturing the complex underlying relationship in many real-world energy forecasting problems \cite{Raza}. 

Recently, with the growing emerging of sensing technology and, therefore, the voluminousness in energy time series datasets, the demand for relevant forecasting models has grown. Accordingly, an ANN with a shallow architecture might not be proper to cope with such voluminousness and complexities in datasets, particularly when attempting to model long interval and nonlinear time series dataset \cite{QZhang}. Consequently, deep learning has emerged in the AI field achieving impressive performance in a vast range of classification and regression applications \cite{Ben, DL}. In energy research, there is a remarkable potential from researchers to use deep learning approaches to solve various energy TSF problems \cite{Shi, Tanveer}.

\subsection {Related work}
It is expected that the growing research interest in developing efficient energy management systems to continue in the light of global sustainability drive \cite{EMS}. This makes real time energy forecasting systems very important in this regard. In line with this direction, several comprehensive reviews are carried out, in the last two decades, addressing the problem of energy forecasting \cite{NanWei}. For example, in 2001, Hippert et al. \cite{Hippert} presented a systematic review on short-term load forecasting using ANNs. After ten years, Zhao et al. \cite {Zhaoh} presented another systematic review for statistical and AI techniques on prediction of building energy consumption. In 2015, Raza et al. \cite{Raza} presented a review on short-term load forecasting based on AI techniques. In the same years, Martinez-Alvarez et al. \cite {Martinez} presented a survey on data mining techniques for time series forecasting of electricity. Daut et al. \cite{Daut} presented a review on the problem of building electrical energy consumption using conventional and AI methods. 

However, a number of recent articles, handling various problems in the energy forecasting domain, are comparable to our paper. For example, in 2017, Deb et al. \cite {Deb} presented a a systematic review for nine popular machine learning techniques for forecasting univariate and multivariate time series energy consumption. No experiments are conducted in their paper, where they relied on the experiments described in other papers that used any of these nine techniques. In 2018, Chou et al. \cite {Chou} presented another review of machine learning techniques using data collected from a smart grid installed in a building. They proposed a number of hybrid models built on platforms, e.g. MATLAB, which are not easy to use. In addition, they base their finding on one short-term dataset, which does not ensure generalization. 

In 2019, Wei et al. \cite {NanWei} presented a similar review of conventional methods and AI-based methods for energy consumption across various forecasting horizons. Wei's review showed that conventional methods outperform the AI-based methods. Like other reviews, no experiments are conducted in their paper, where they reviewed the techniques described in other articles. In the same year, Divina et al. \cite {Divina} presented a comparative study of different forecasting methods for energy consumption of smart buildings. Divina's study showed that the machine learning methods are the ideal to achieve this task.

Certainly, the aforementioned reviews provide vital information about various energy forecasting strategies through different scales and horizons. Nevertheless, some of the earlier related works are based on static data, which usually fits a dependent variable to a set of independent variables \cite {Hippert,Zhaoh}. Most of the recent reported reviews on the energy forecasting domain are systematic reviews in a specific energy problem. A systematic research, sometimes called a taxonomy research, refers to the process of systematically dividing the baseline problem into several categories \cite{HWang}. The author of a systematic review, or taxonomy, paper usually counts the number of published articles in each category, for example, a review for 40 papers in \cite {Hippert}, 116 papers in \cite{NanWei}, and 157 papers in \cite{Deb}.  

Yet, the most common practice in most of these taxonomy articles is comparing the performance of existing methods on different datasets and different experimental conditions; for example \cite{Raza, Daut, Deb, NanWei}. In such cases, the evaluation of existing methods has not a unified base of comparison, which therefore reduces such review's benefit. There are very few review papers that conducted an empirical comparison besides the analytical review but using one dataset, such as \cite{Divina, Chou}.    

\subsection {Objectives of the paper}

To fill the aforementioned research gaps, the objective of this study is twofold. \textbf{First}, it presents an analytical study for 15 common varied techniques that can be used to solve the energy TSF problems. These techniques distributed among conventional techniques, machine learning techniques, and deep learning techniques. The analytical part describes each technique inside the paper and ends with a basis of qualitative comparison among these 15 techniques listed in a Table \ref{Overall_Comparison} at the end of this paper. The aim of this qualitative comparison is to understand these techniques deeply from the energy forecasting perspective, and to identify their advantages and disadvantages. \textbf{Second}, the paper presents a quantitative empirical assessment by comparing the performance of all techniques in solving three different energy TSF problems. To ensure a fair assessment, the experiments carried out in a unified experimental conditions and using three public datasets, with different sizes. The datasets mix between short-term and long-term horizons. 

Accordingly, we can summarize the objectives (or contributions) of the paper in the following items:
\begin{enumerate}
\item Provide a comprehensive review of the energy TSF problems along with a review to 15 most common techniques.
\item Carry out a qualitative analysis of these techniques identifies the advantages and disadvantages.
\item Carry out a quantitative empirical comparison among these techniques in terms of accuracies and computational requirements.
\item Assess the performance of deep learning techniques to solve various energy TSF problems.

\end{enumerate}

Finally, it is worth mentioning that this study focuses on the optimization of conventional, machine learning, and deep learning methods that fit to solve the energy TSF problem, rather than looking into the optimization aspects of the energy consumption or demand. Consequently, the scope of this study can be easily extended into other TSF domains. 

The remaining of the paper is organized as follows. Section 2 shows a number of basic concepts. Section 3 shows a description of three categories of techniques that used to solve the TSF problem along with a brief overview of selected models from each category. Section 4 describes the forecasting evaluation metrics that used to assess the forecasting techniques. The dataset configuration and the empirical results for each case study are provided in section 5. Section 6 demonstrates a discussion and analysis of the earned results. A number of challenges and future research directions are given in section 7. Section 8 shows the paper conclusion and recommendations.

\section{Background}
This section shows a few basic concepts that will help the reader to go through the paper. 

\subsection {Statement of a TSF Problem}
A time series is a sequence set of observations or data points measured in a proper chronological order of a variable of interest. It can be either a discrete series or a continuous series. In the discrete time series, the observations are measured at discrete time interval, whereas in the continuous time series, the observations are measured at every time instance. If a time series contains observations of one variable, it is denoted as a univariate time series (UTS), otherwise it is a multivariate time series (MTS) \cite{Wei}.

A TSF problem is the task of predicting future values of time series data either using previous data of the same signal, i.e. UTS forecasting, or using previous data of other correlated signals, i.e. MTS forecasting. Mathematically, the UTS problem can be formalized as a sequence of $n$ real-valued numbers $x = \{ x(i) \in \mathcal{R}:i=1,2,...,n \}$ where $n$ represents the series length that recorded for the corresponding pattern. 

In UTS forecasting problems, the forecasting model usually predicts the variable of interest $x$ using past values that precede $x$. In this manner, the forecasting model perceives the structure and learns the trend of the underlying pattern and extrapolates the interested process into the future. On the other hand, the MTS forecasting problem has two or more correlated variables. It can be defined as a finite sequence of several UTS problems, such that each UTS problem models a different pattern. Then the following formula,

\begin{eqnarray}
 X = (x_1, x_2, ..., x_m)
\end{eqnarray}

\noindent represents an MTS problem includes $m$ variables, where the corresponding component of the $j$th variable $x_j$ is a UTS problem of length $n$ and can be given as,

\begin{equation} \label{UTS}
x = \{x_j(i) \in \mathcal{R}:i=1,2,..,n\}(j = 1,2,..,m)\end{equation}

Regardless of a univariate or a multivariate, the common TSF problems are divided into three categories based on the time horizon \cite{Douglas}, as follows,
\begin{enumerate}
\item Short-term forecasting problem that extends from one hour to one day or weeks or months ahead.
\item Medium term forecasting problem that extends from one month to one or two years ahead
\item Long-term forecasting problem that extends from one year up to ten years, or more,  ahead.
\end{enumerate}

It is demonstrated that, there are four factors that affect the observations of a time series, as follows, where the initial two factors are the most challenging to a forecasting model \cite{Douglas}.
\begin{enumerate}
\item Seasonality - They are patterns of short-term repeated change in a time series during the same time every period. For example,  the power consumption in summer days is generally high compared to those in winter days through a year. 
\item Trend - It is the smooth long-term direction of a time series of increasing or decreasing patterns during a very long period, like electricity price. 
\item Cyclic - It is a pattern of the rise and fall of a time series over periods longer than one year, and it depends on the type of problem.
\item Irregularity - It is the residual of the time series after removing all of the seasonal, trend, and cyclic components.
\end{enumerate}

\subsection{Energy TSF Problem} 
The energy TSF problem is a challenging problem due to significant volatility and uncertainty of natural factors that included. It is widely demonstrated that energy time series datasets are complex and has an abnormal distribution \cite{Dannecker}. In addition, they have high nonlinearity and nonstationarity characteristics. These limitations make energy time series data are difficult to be analysed using either statistical or conventional computing methods. Furthermore, due to the economic and population growth, the consumption of energy resources had increased dramatically during recent years. 

Accordingly, the development in energy TSF has significant research values and should attract extra research efforts in order to mitigate any expected energy crisis in the future \cite{Dannecker, Provo}. In this paper, three different energy TSF problems are examined. In the following, a brief description for each problem is given.

\subsubsection{Electrical energy consumption problem}
The electrical energy consumption and its forecasting techniques are very crucial to stakeholders for estimating the electric energy usage as well as making right decisions for future development to expand power systems. Energy stakeholders use energy consumption forecasting models to monitor the change of energy consumption attitude and compared the outcomes with the predicted values over a certain period. For this reason, many forecasting algorithms focusing on electrical energy consumption are presented for the sake of improving the electrical energy efficiency as indicated in \cite{Deb,chu, Zagrebina, zhuang,rahman, voyant}. In fact, there are several factors, which make electrical energy consumption forecasting a challenging problem, which can be summarized as follows.
\begin{itemize}
\item {Big data challenge:} The problem of voluminous and dimensionality of the time series data arises and, consequently, time series analysis based on learning-based techniques are required.

\item{Change in consumers consumption behavior:} The energy consumption has a dynamic behavior and may change every minute based on changes in consumers activities and events. 

\item{Weather conditions:} Energy consumption can be vary depending on the state of the weather.

\item{Demography:} Rate of population growth is not fixed every year and it is in increasing pattern and, hence, the energy consumption. 

\end{itemize}

\subsubsection{Natural gas consumption problem}
In contrast to the electrical energy field, the prediction activities in the natural gas field is still immature \cite{jolanta}, however the significant importance of natural gas for human and industry. As confirmed by the International Energy Outlook \cite{out16}, the natural gas will remain an essential resource until, at least, 2040. The production and delivery processes of natural gas pass by three main phases; namely, production, transmission and storage, and distribution \cite{eia}. The consumption prediction is also an important process in this industry. Similar to electrical energy consumption, the variability of natural gas consumption over time depends on similar external factors \cite{jolanta}.

\subsubsection{Household load forecasting problem}
Individual household load forecasting is another challenging problem in the energy domain in terms of system volatility due to dynamic attitude composed of many individual components \cite{Shi}. This problem is influenced by a number of external factors, including customers attitude, devices characteristics, time and day of the week, holidays, geographic patterns, weather conditions, and other economical factors. In the recent years, the problem of household load forecasting received much attention after the emerging of smart grids and the advent of advanced metering infrastructure, energy storage systems, and home area networks.

\subsection {Parametric vs. nonparametric techniques}

In sensory-based applications, one of the key requirements is to maintain integrity of the underlying sensory data; so that it can be monitored, analyzed, and managed in a trusted manner. Sensory data integrity has been analyzed by either parametric or nonparametric techniques. Parametric method is a learning model that can represent the data in a fixed size set of parameters regardless of number of training samples \cite{russell}. Though these techniques are simple and fast to learn the features included in the data using a functional form, they are highly constrained to this specified form. In addition, regression models based on parametric techniques are highly sensitive to the models' parameters as well as they are unlikely to match the underlying mapping function. Examples of these techniques are Bayesian, Naive Bayes, Perceptron, linear discriminant analysis, and logistic regression.

Unlike parametric techniques, the nonparametric techniques do not assume a specific formula for the learning function. In other words, there are no restrictions to learn any function using the training data. Nonparametric techniques are very suitable when there is no prior knowledge about the learning function as well as there are enough training data to construct the mapping function \cite{russell}. Accordingly, these techniques are maintaining some ability to generalize to testing data and much flexibility to fit a large number of functional forms. Despite these advantages, these techniques require plentiful of training data to be able to estimate the functional form, which will cause the model's slowness as well as overfitting. Examples of these techniques are kNN, ANNs, decision trees, and SVM. The following section shows different models, which fall down either parametric or nonparametric. 

\section{Methodologies}

\begin{figure}[h!]
\begin{center}
  \includegraphics[width=0.5\textwidth,height=14pc]{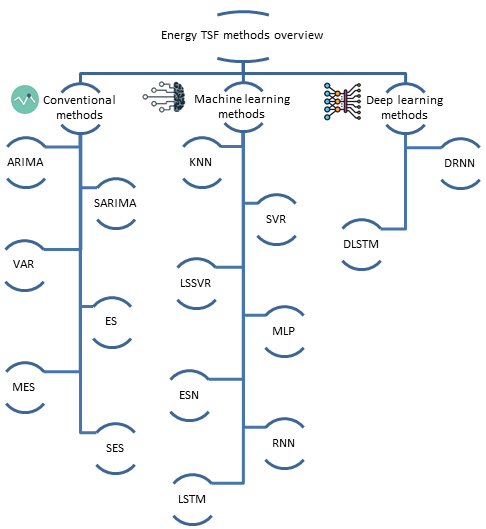}
  \end{center}
  \caption{Stream of selected techniques}\label{Chart}
\end{figure}

This section shows a brief description of a number of conventional, machine learning, and deep learning techniques used in our empirical assessment. Figure \ref{Chart} shows the name and category of selected techniques that covered in this paper. These techniques have selected specifically in terms of their astonishing performance in solving TSF problems. For each of these techniques, we can find myriads of variations developed in the literature. Beside the description given in this section, Table \ref{Overall_Comparison} at the end of this paper concludes this description focusing on advantages and disadvantages of each technique.

\subsection{Conventional methods}
Conventional methods use varieties of mathematical statistics, probability theory, and stochastic processes to establish a mapping between historical time series data and the generated outputs.
Generally speaking, conventional methods have a simple modeling process, when compared with other methods. In addition, they require fewer input data, compared to learning-based methods, and show good performance particularly in the yearly energy consumption forecasting \cite{NanWei}. Though several conventional methods exist, ARIMA, VAR, and the exponential smoothing (SE) methods are the most commonly used conventional methods can to achieve energy forecasting. 

\subsubsection{Autoregressive integrated moving average}
Previously, most of the conventional methods were based on the autoregression (AR) and the moving average (MA) methods, which are accomplished by the efforts of Yule, Slutsky, Walker, and Yaglom during 1920s \cite{rob}. Shortly after, Wold combined both AR and MA methods to be as one method known as ARMA, which can be used to model all stationary time series data \cite{wold}. Precisely, as a parametric model, both the mean of the series and the covariance among observations do not change with time \cite {Siroos}. 

Next, Box and Jenkins popularized the use of the ARMA method and propose an integrated model known as ARIMA, which stands for autoregressive integrated moving average \cite{Box}. Since then, the ARIMA model is widely employed in the forecasting activities as it can be used to solve nonstationary time series problems. In addition, the ARIMA model is highly efficient and widely used in short-term forecasting problems as short-term factors are expected to change slowly \cite{Levenbach}. The ARIMA modeling for a time series can be represented as a linear relation as follows \cite{Deb},

\begin{equation}
\label{arima}
\hat{y}= y + e
\end{equation} 
where \textit{y} is the output of a time series prediction, \textit{y} is the weighted sum of past values, whereas \textit{e} is the error series \cite{Box}. In a mathematical description, we can represent the ARIMA model using three parameters; namely, \textit{p}, \textit{d}, and \textit{q}, as follows,

\begin{equation}\label{ARIMApddq}
ARIMA(p,d,q)= AR(p)+I(d)+MA(q)
\end{equation}
where \textit{p} and \textit{q} represent the  order  of  autoregressive  term and the order of moving average term, respectively.  The parameter \textit{d} refers to number of times the series is differenced  in  order  to  make  the series more  stationary.  If  it is already stationary, then no differencing is required; i.e. d~=~0, and in this case it is known as ARMA. \textit{AR} (\textit{p}) represents $y_t$; the value to  be  predicted  considering the previous \textit{p} values of y and given as,

\begin{eqnarray}\label{eq6}
y_t=\sum_{i=1}^{p}  \phi_{i}~  y_{t-i}+e_t
\end{eqnarray}
while, \textit{MA} (\textit{q}) represents $y_t$; the value to be predicted considering the previous \textit{\textit{q}} error values and given as,

\begin{eqnarray}\label{eq7}
y_t=\sum_{j=1}^{q}  e_{j}~  y_{t-j}+e_t
\end{eqnarray}
Thus, the full mathematical formula of ARIMA represented in Eq. (\ref{arima}) can be given as,
 
 \begin{eqnarray} \label{fullarima}
 y_t= c+ \sum_{i=1}^{p}  \phi_{i}~  y_{t-i}+\sum_{j=1}^{q}  e_{j}~  y_{t-j}+e_t
 \end{eqnarray}
 
However, several ARIMAs' variants were discussed as special cases of the original ARIMA model, such as white noise, autoregression, and random walk w/o drift \cite{rob}. So,  it is much easier to work with the back-shift notation (\textit{B}) to describe the process of differencing and combining the components (i.e. AR, I, and MA) to form the complicated and special cases of the ARIMA model. Thus, Eq. \ref{fullarima} can be rewritten in the back-shift notation as \cite{rob},
 
  \begin{eqnarray} \label{backshift}
AR(q)  \quad \quad \quad \quad I(d)  ~~~~~~~~~~~~~ \nonumber \\
    \downarrow ~~~~~~~~~~~~~~~~~~~~~~~~~ \downarrow ~~~~~~~~~~~~~~~~ \nonumber \\ 
(1-\phi_{1}B- .. - \phi_p B^p) (1-B)^d y_t =  \nonumber \\
 c +(1+ \theta_1B+... +\theta_q B^q)\epsilon_t \\
 \uparrow ~~~~~~~~~~~~~~~~~~~~~~\nonumber \\
 MA(q) ~~~~~~~~~~~~~~~~~ \nonumber
 \end{eqnarray}
 
Many articles were published in the literature addressing the ARMA and ARIMA models to solve energy problems. For example, Chujai et al. \cite{chu} used the ARMA and ARIMA models in forecasting the individual household electric power consumption. Rehman et al. \cite{Rehman} applied ARIMA to analyse the energy demand up to 2035 for electricity, oil, natural gas, and coal in Pakistan across all sectors. They found that ARIMA outperformed other models using annual energy demand data. Ediger et al. \cite{Ediger} used ARIMA to predict the levels of natural gas energy demand in Turkey.

However, ARIMA has a number of limitations such as the disability to represent dynamic behaviors and the non-linearity of energy time series data \cite{Deb, mar}. In addition, ARIMA does not support the seasonality in the energy time series data. To solve the seasonality problem, an extension of ARIMA is developed called seasonal ARIMA (SARIMA) \cite{chang}. Nevertheless, this extension adds three new hyper-parameters to ARIMA specify the seasonal component of the series, as well as an additional parameter represents the number of periods in each season.
It can be represented as $ARIMA(p,d,q)(P,D,Q)_{m}$, where (\textit{p},~\textit{d},~\textit{q}) for non seasonal part and (\textit{P},~\textit{D},~\textit{Q}) for seasonal part with (\textit{m}) seasonal period \cite{rob, chang}. Simply, we can say that the additional seasonal terms are multiplied by the non-seasonal terms. However efficient prediction using univariate time series data, ARIMA is not suitable to do the same with a multivariate time series (MTS) data. This is due to the disability to represent the dynamic behavior of a multi variable dataset.  

\subsubsection{Vector autoregression (VAR)}
One of the commonly used generalization methods of ARIMA is the VAR method, which provides a flexible means to model and forecast MTS problems \cite{Lutk}. VAR is a parametric model and each variable is a linear function of the past values of itself and the past values of all the other variables \cite{Kali}. In addition, VAR enable the user to measure and visualise the estimated effect of the explanatory change on the dependent variable over time \cite{Lutk}. 

The VAR model that containing \textit{n} time series variables and a lag length of \textit{k} is written as,

\begin{equation}
y_t=\alpha_0+\alpha_1y_{t-1}+\alpha_2y_{t-2}+...+\alpha_ny_{t-n}+u_t
\end{equation}

\noindent where $\alpha_0$ is an (\textit{n} x 1) vector of intercepts, $\alpha_i$: $i \leq 0$, are (\textit{n} x \textit{n}) coefficient matrices, $y_t$ is an (\textit{n} x 1) vector of endogenous variables, and $u_t$ is an (\textit{n} x 1) vector of white noise residuals. The vector $\alpha_0$ contains \textit{n} intercept terms and each matrix $\alpha_i$ contains $n^2$ coefficients. Therefore, number of overall coefficients that must be estimated is $(n+kn^2)$, which grows up exponentially with the number of variables in the system. Accordingly, a major problem in the computation of the VAR model occurs when \textit{k} is large, which yields over parameterization. As a result, too many coefficients must be estimated in proportional to the sample size \cite{Lutk}. 

A variant VAR model, called Bayesian VAR (BVAR), could avoid the expensive computation of the original VAR by allowing the model to include many coefficients while simultaneously controlling their influence by the data. This improves the forecast performance by reducing the number of false correlations that captured by the original VAR model \cite{Kali}. It is very recently when VAR used to make an energy forecasting using univariate and multivariate datasets, where VAR outperforms other autoregressive methods \cite{shah}. For example, Liu et al. \cite{Liu} built a VAR forecast model based on three weather variables, for 61 cities in the US, to model the electricity supply and demand.

\subsubsection{Exponential smoothing (ES)}
The ES is a traditional statistical parametric method, introduced in the 1950s, commonly used in solving TSF problems. It is considered as a collection of ad-hoc techniques for extrapolating various types of UTS data. Compared to ARIMA, the main concept of ES depends upon the assumption of exponential decay of weights for past data over time while ARIMA is employed by converting the time series into a stationary series \cite{Oliv}. 

Though, ES has a solid theoretical basis and works well even with fewer input data, it suffers when the data has a trend or a seasonality particularly with long-term data. In addition and similar to ARIMA, it is not suitable to model the nonlinearity of MTS data. There is a simple version of the ES method called simple exponential smoothing (SES). The SES method is used only when there is no clear trend or seasonal patterns in the data.  Eq. (\ref{seseq}) shows the degrade of the weights exponentially, where $\alpha$ $(0 \leq \alpha \leq 1)$ determines at which rate they are decreasing.

\begin{eqnarray}
\label{seseq}
\hat{y}_{t+1|t}&=& \alpha y_t+\alpha(\alpha-1)y_{t-1} \nonumber\\
&&+ \alpha(\alpha-1)^2y_{t-2}+ ...
\end{eqnarray}

Holt extended the SES method to allow the forecasting of the data that include trend \cite{holt}. Also, Holt and Winters extended the Holt’s method to capture seasonality, which includes a seasonality smoothing parameter $\gamma$, by modeling three types of exponential smoothing: a value, a trend over time, and a seasonality pattern \cite{rob}. Independently of the univariate ES, the multivariate form of this method has developed in the form of a state space model. One of the early contributions that outlined a multivariate ES (MES) seasonal specification was presented by Pfeffermann et al. using two univariate tourist data sets \cite {pfeff}. In their paper, the authors compared MES method with the VAR method and the original ES method. The experimental results showed that MES produced more accurate results than other contenders.

\subsubsection{Epilog}
Overall, conventional forecasting methodologies have drawn much attention due to their relative simplicity in representation as well as relying on solid theoretical and mathematical bases. However, most of the conventional methods are linear and parametric and, therefore, show a poor forecasting performance, particularly when treat long-term data and MTS problems \cite{Zhang-Book}. In addition, the variables in most of the energy applications exhibit a highly nonlinear attitude \cite{Levenbach}. Therefore, the use of conventional methodologies to model these nonlinear variables is not appropriate because of the poor representation of the complex relations and interactions among energy variables \cite{Carlos}. 

\subsection{Machine learning methods}
Machine learning methods have flexible structures and nonparametric procedures sufficient for capturing and identifying any complex interactions and nonlinearity relationship among the variables of the TSF problem \cite{Parmezan}. Simply, machine learning defined as the automated frame of the human being learning. It is demonstrated that human beings learn through experience using a trial and error style in order to discover which actions should be triggered given certain circumstances. This enables the human to make abstractions and build knowledge. The machine learning is similar; it can be regarded as the algorithms that have the objective to improve a performance measure by automatically extracting its own rules and creating its own models based on the given information \cite{mar}. 

During the last two decades, there has been a growing interest in employing various machine learning models in the forecasting domain, particularly the energy TSF problem \cite {voyant}. The most important aspect of machine learning methodologies versus conventional methodologies is their capability to accomplish the learning process in order to improve the model performance over time using a trial and error fashion \cite{mar}. Artificial neural networks (ANNs) are the most machine learning algorithms widely used in the energy forecasting applications. 

\subsubsection{Artificial neural network (ANN)}
\begin{figure}[h!]
\centering
\includegraphics[width=0.45\textwidth,height=10pc]{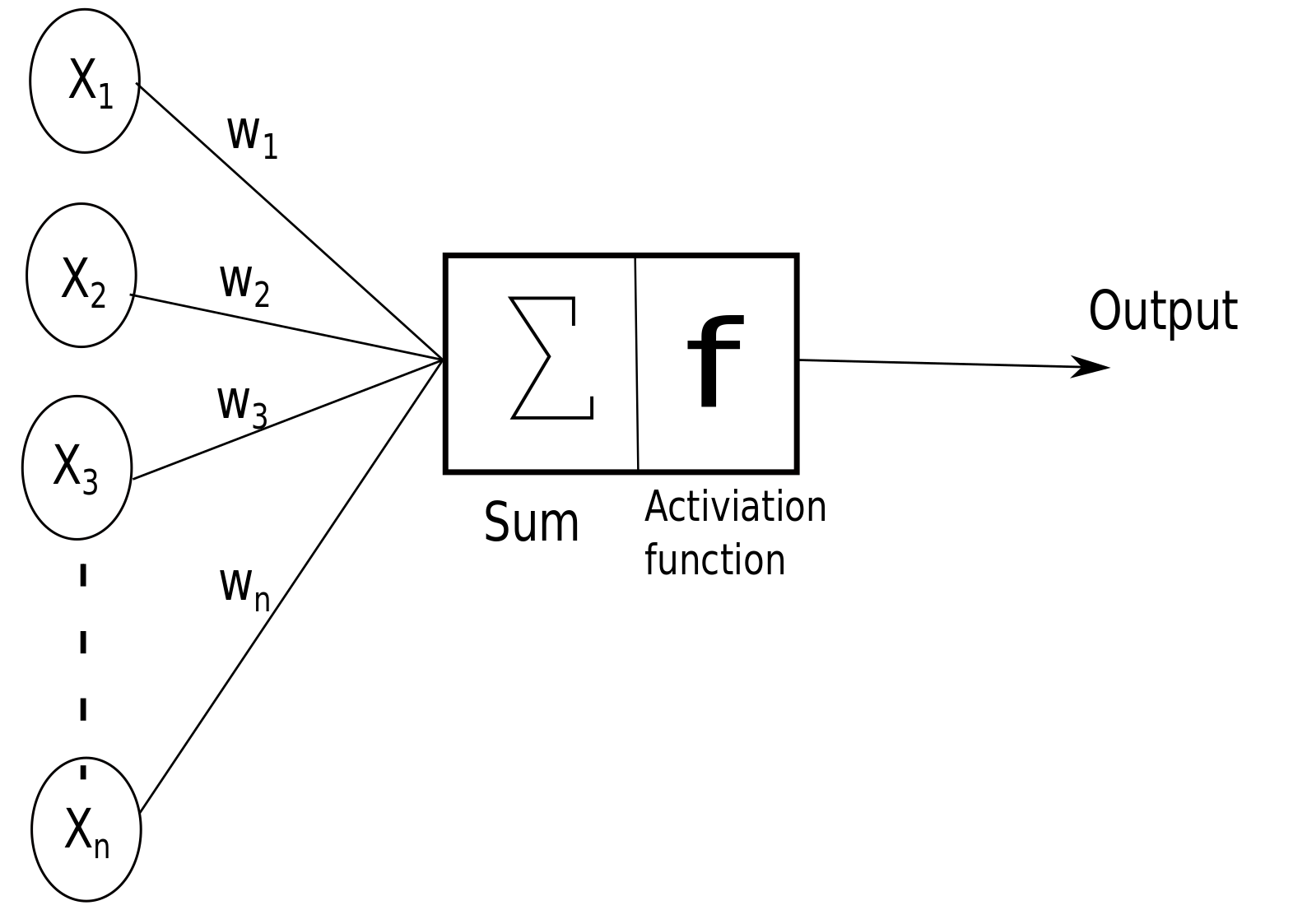}
\caption{The standard artificial neural network architecture.}\label{ANN}
\end{figure}

ANN is a modeling technique that mimics the functions of the biological neural network in the human brain. As shown in Figure \ref{ANN}, ANN consists of many computing network units linked by directed connections capable of performing a lot of complex computations. The network outputs are given through all outgoing connections, where $w_i$ represents the connection's strength with the input $x_i$. The ANNs algorithms are decent enough for dealing with the intrinsic properties commonly exist in the energy activities \cite{Dannecker}. For more details about ANNs and forecasting see \cite{Zhang-Book, Tealab}.   

Many advantages are reported for ANNs such as their ability to efficiently deal with the extreme noisy data and prone to error time series. In addition, ANNs have the ability to manipulate the complexity and nonlinearity of variables and processes in the energy domain; the problem that can be alleviated by the amenable mathematical ANNs structure \cite{Carlos}. Moreover, ANNs are considered as a parallel distributed processor which is capable of storing information for further use. 

Nevertheless, the most important aspect in the treatment of the TSF problems using ANNs is that the statistical distribution of the raw time series data need not be known in advance. This is due to that the nonstationarities aspects, which exist in the time series data, such as trends and seasonality are implicitly estimated by the internal structure of the ANN \cite{Carlos}. Hence, machine learning and ANNs algorithms are deemed powerful alternatives to conventional methods. In the following subsections, we will show a brief description of the most known machine learning and ANNs algorithms that can be employed to solve various energy TSF problems.

\subsubsection{k-nearest neighbour (kNN)}
The kNN algorithm is a nonparametric method used for both classification and regression problems \cite{kNN}. In both cases, the input consists of the \textit{k} closest training examples in the feature space. The output depends on whether kNN is used for classification or regression:
\begin{itemize}
\item In kNN classification, the output is the class's membership. An object is classified by a majority vote of its neighbours, with the object being assigned to the most common class among its \textit{k} nearest neighbours (\textit{k} is a positive small integer). If \textit{k} = 1, then the object is simply assigned to the class with a single nearest neighbour.
\item In kNN regression, the output is the property value for the object. This value is the average of the values of its \textit{k} nearest neighbours.
\end{itemize}

In case of regression, given a data point, we compute the Euclidean distance between this point and all points in the training set. Then, picking the closest \textit{k} training data points and set the prediction as the average of the target output values for these \textit{k} points. Considering \textit{J}(\textit{x}) is the set of \textit{k} nearest neighbours of point \textit{x}, then the prediction y is given by,
\begin{eqnarray}
y= \sum_{m  \in J(x)} \frac {y_m}{k}
\end{eqnarray}
where $y_m$ is the target output for training data point $x_m$. Note that, large kNN will lead to a smoother fitting and lower variance, and vice versa for a small \textit{k}.

In the field of energy, Lora et al. \cite{lora} proposed a method based on the kNN to solve the TSF problem applied to short-term electric load forecasting. The empirical results lead up to the kNN method is more accurate than dynamic regression models to solve the selected problem. Sun et al. \cite{sun} suggested a kNN based technique for decreasing the cost of energy that home energy management system is using. kNN helped the authors to analyze the classification and regression datasets they have and to make a simulation of the energy needed for each apparatus. Johannesen et al. \cite{johannesen} developed a regression approach using kNN to explore the use of regression of regional electric load forecasting by correlating lower distinctive categorical level (season, day of the week) and weather parameters.

Nevertheless, employing kNN in forecasting applications is limited due to some limitations \cite{Imandoust}. Specifically, kNN may show a poor run-time performance when the training dataset is large. In such a case, computation cost will be quite high since we need to compute the distance of each query instance to all training samples. In addition, it is very sensitive to irrelevant or redundant features because all features contribute to the similarity and thus to the classification or regression process, however, with a careful feature selection (or weighting) this limitation can be avoided. Moreover, there is a limitation is related to distance metric, where it is not clear which type of distance metric and attributes to use to yield the best results \cite{Pan}.

\subsubsection{Support vector regression (SVR)}
The support vector machine (SVM) is a machine learning algorithm widely used in classification and recognition applications \cite{Vap2}. When SVM is applied to achieve regression analysis of TSF problems, it is called as support vector regression (SVR) \cite{SVM}. SVM is based on two concepts of statistical learning theory; namely, the decision plane and the decision boundary. The decision plane can be defined as a plane that separates a set of different objects. Basically, the SVM  uses a linear function to implement nonlinear class boundaries through a nonlinear mapping of the input vectors \textit{x} into a high-dimensional feature space \cite{Vap2}.

\begin{figure}
\begin{center}
  \includegraphics[width=0.5\textwidth,height=14pc]{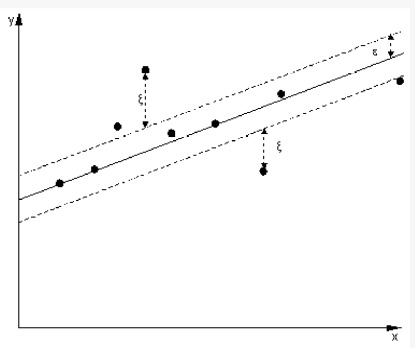}
  \end{center}
  \caption{The epsilon-insensitive band via one-dimensional linear regression function} \label{svm}
\end{figure}

Similar to SVM approach, there is motivation to seek and optimize the generalization bounds given for regression via SVR. Actually, SVR depends on defining the loss function with ignoring the possible errors that are located within a certain distance from the actual values. This loss function is denoted as epsilon-insensitive loss function, as shown in Figure \ref{svm}, which embeds a one-dimensional linear regression function with epsilon-insensitive band. The goal of SVR is to find a function \textit{f}(\textit{x}) = \textit{$w^T$}\textit{x} + \textit{b} that
deviates no more than $\zeta$ from the targets $y_i$ for all training data. 

\begin{figure}
\begin{center}
  \includegraphics[width=0.45\textwidth,height=14pc]{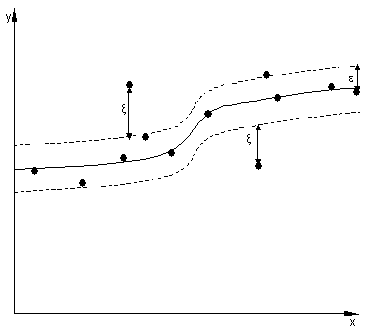}
  \end{center}
  \caption{The epsilon-insensitive band via one-dimensional nonlinear regression function} \label{svm2}
\end{figure}

For linearly separable data, the corresponding quadratic optimization problem is given as,
\begin{eqnarray}\label{eq8}
&Minimize &~ \frac{1}{2} w^T w \nonumber\\
&subject~ to& ~ y_i (w^T x_i + b) \geq  1; \forall i = 1, 2,..., N
\end{eqnarray}
where \textit{w} is the weight vector, \textit{b} is the bias. The SVM can deal with nonlinear regression cases as shown in Figure \ref{svm2}. For the nonlinearly separable data, the corresponding optimization problem can be given as,
\begin{eqnarray}\label{eq9}
&Minimize &~ \frac{1}{2} w^T w+C{\sum^{N}_{i=1} \zeta} \nonumber\\
&subject~ to & ~ y_i (w^T x_i + b) \geq {1}-\zeta; \forall i = 1, 2,..., N \nonumber\\             &     \zeta \geq {1} &
\end{eqnarray}
where \textit{C} is a constant.
\\

Suykens et al. developed an extension for SVR and called it as least squares SVR (LSSVR). It represents a higher stability alternative to SVR as well as it is trained faster than SVR \cite{Suykens}. The LSSVR algorithm improves the SVR method by transforming the quadratic programming problem of SVM into a linear equation by establishing a quadratic loss function instead of the epsilon-insensitive loss function. Accordingly, this improves the accuracy along with reduces the computational burden of SVR. Therefore, we can obtain the LSSVR regression model by solving the following optimization problem,

\noindent where $\gamma$ is a constant similar to \textit{C} in the standard SVR, ($x_i$) is the mapping to the high dimensional feature space as in SVR, and \(e_i \in {\textit{R}}\) are the error variables. Even though, the performance of LSSVR degrades if the time series data have chaotic characteristics \cite{Du}. 

\subsubsection{Multilayer perceptron (MLP)}
\begin{figure}[h!]
\begin{center}
  \includegraphics[width=0.45\textwidth,height=14pc]{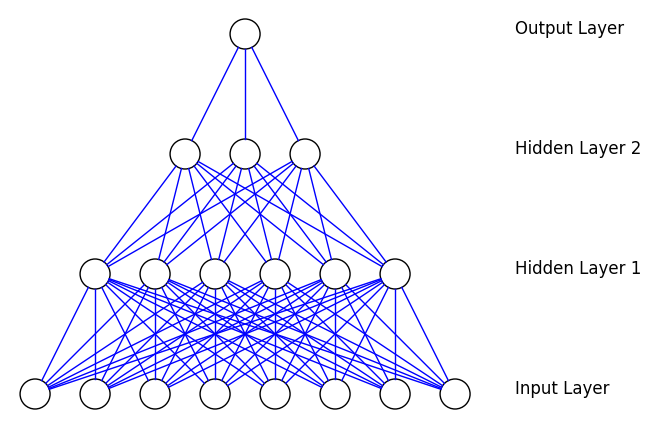}
  \end{center}
  \caption{The multilayer perceptron architecture} \label{mlpp}
\end{figure}

The MLP is a conventional neural network model \cite{Olawo} consists of one input layer, one (or more) hidden layers, and one output layer, as shown in Figure \ref{mlpp}. Each hidden layer includes a number of units, called neurons, that can be considered as a single output perceptron network. The output unit is equivalent to a single output unit perceptron, and it regarded as a soft thresholded linear combination of the units of the preceding hidden layers. The hidden and output units are based on a sigmoid that calculates a linear combination of its input \textit{x}, and then applies the following sigmoid function on the net result,
\begin{equation}\label{mlp}
sigmoid(x)=\frac{1}{1+e^{-x}}
\end{equation}

The learning of MLP employs the backpropagation (BP) algorithm in order to minimize the sum of squared errors \footnote{Both the ANNs and the conventional models keen to improve the forecasting accuracy via minimizing the sum of squared errors, but they are different in how this minimization process is carried out.} by changing the connection weights among neurons. Ideally, the MLP aims to construct a model that is capable of mapping the inputs to the known outputs using previous historical data. The produced model can be used to predict unknown outputs. 

In TSF terminology, the number of neurons in the input layer of the MLP network is equivalent to the number of independent variables such as the number of days, day of the month, number of hours. While the number of neurons in the output layer is equivalent to the number of dependent variables such as the amount of gas or power consumption. In addition, the MLP model will learn the function from lag observations of time series data \cite{Brown}. Due to this simple representation, the MLP is commonly used in power, load and gas forecasting problems \cite{vv,Voyant2}. In the energy time series data, MLP can learn from multiple observations that taken through prior time steps, or lag observations, and use them as input features. In addition, MLP has the ability to use these input features to achieve multiple-step ahead forecasts \cite{Brown}.  

However, MLP has some drawbacks like slow convergence, linearity, a tendency to get trapped into local minima, possible oscillations during searching and sensitivity to learning rate \cite{serkan}. In addition, determining the best MLP architecture requires making a large number of experiments, which make them impractical in terms of the data voluminous. Moreover, increasing the number of layers and neurons yield overfitting and training difficulties \cite {Tien}.
 
 \subsubsection{Recurrent neural network (RNN)}
\begin{figure}[h!]
\begin{center}
  \includegraphics[width=0.4\textwidth,height=9pc]{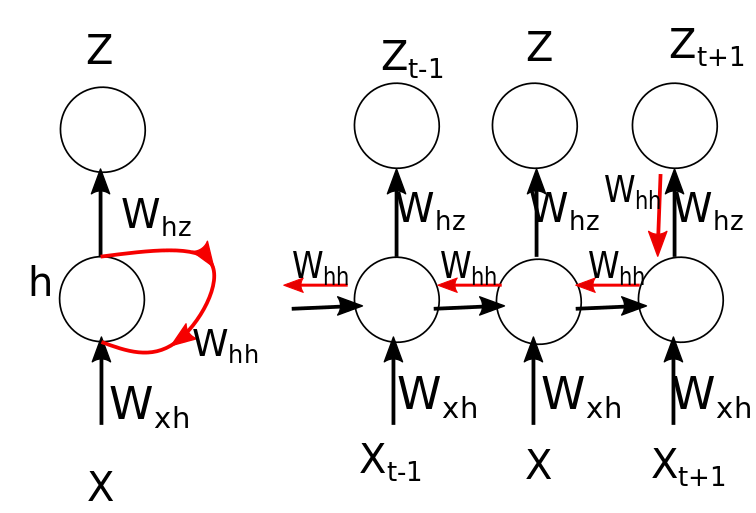}
  \end{center}
  \caption{RNN Architecture, h refers to the hidden state, X refers to the observation, W refers to the cell's weight, and Z refers to the output} \label{RNN}
\end{figure}

The RNN is a widespread neural network approach adopted to solve the energy TSF problems. Compared to other traditional neural networks, the RNN has a looping mechanism that allowing information to flow from one step to the next step as shown in Figure \ref{RNN}. This information is the hidden state, which is a representation of previous inputs \cite{Lipton}. The hidden state $h_{t}$ is a nonlinear mapping depends on the current input $x_t$ and the previous hidden state $h_{t-1}$. It is written as,
\begin{eqnarray}
h_t = f (h_{t-1}, x_t)
\end{eqnarray}

As it depicted in Figure \ref{RNN}, the structure of standard RNN contains an internal memory cell. It computes recursively  a new output by applying an activation function into the previous historical data and new inputs. This allows the RNNs to process information sequentially and exhibits temporal behaviour for a time sequence while retaining information from the past \cite{Lipton}. As a result of this distinct structure, many attempts have been performed to use RNNs in power, electric load and natural gas forecasting domains \cite{Zagrebina, Omer, Bing}. 

Nevertheless, in the context of TSF, the main drawback of RNNs is that they require essentially intensive connections among cells, as well as much memory in simulation than the other BP-based neural networks. In addition, RNNs are not able to keep track of long-term dependencies because of the vanishing and exploding gradient problems, which prevent the information from propagating to early layers in the architecture \cite{jiong}. Several remedies have been developed to address the drawbacks of RNN, the long short-term memory and the echo state network. 

\subsubsection{Long short-term memory (LSTM)}
When learning a sequential data, the standard RNN aims to learn representations of patterns repeatedly occurred via the past observation by sharing parameters across all time steps. But as the time goes on, the memory of past learned patterns is fade. As a special type of RNN but with a different structure, LSTM allows the model memory cell to memorize the data sequence for a longer period of time by establishing propagation tracks keep the flow of gradients for earlier states \cite{kong}.

\begin{figure}[h!]
\begin{center}
  \includegraphics[width=0.5\textwidth,height=11pc]{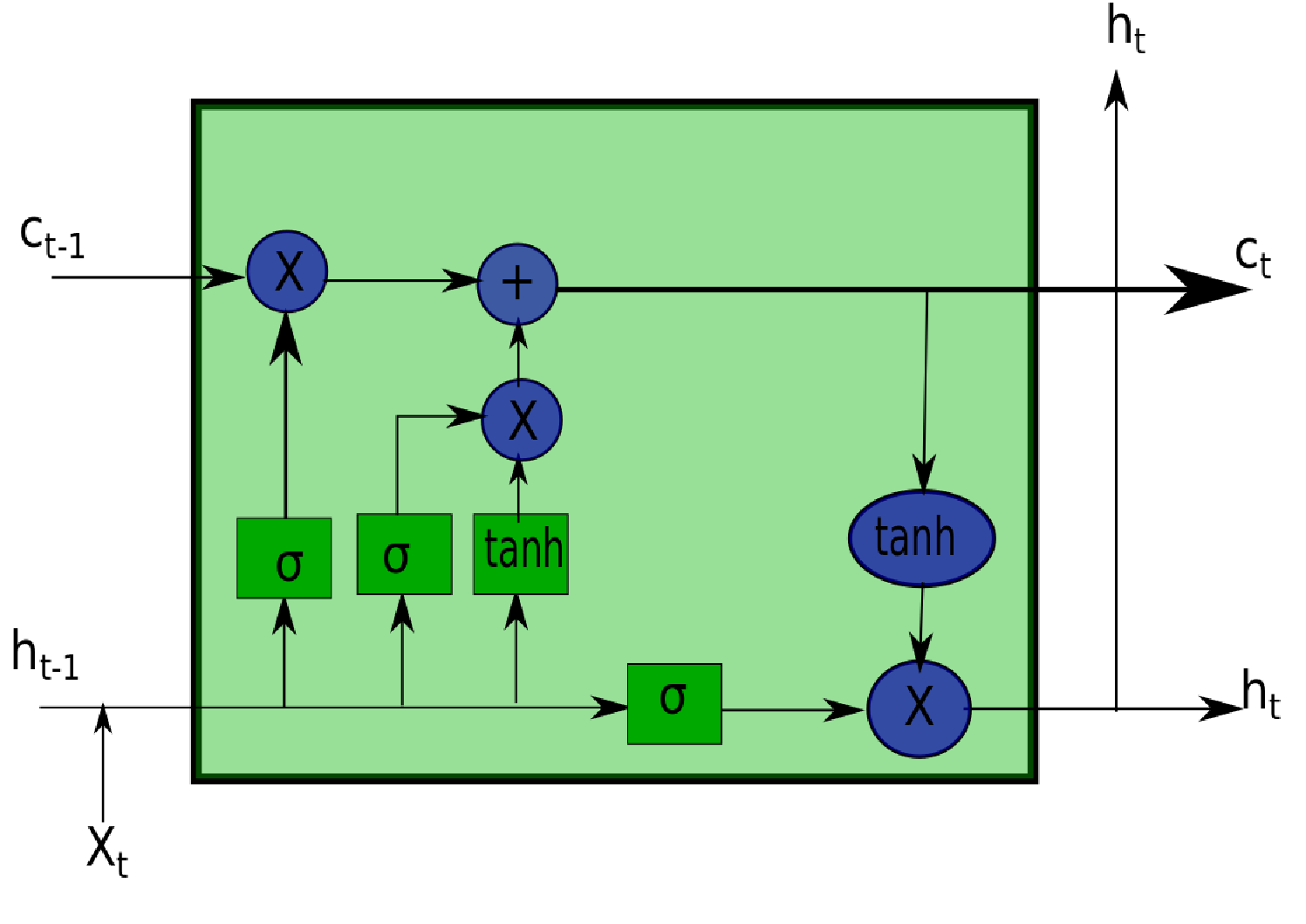}
  \end{center}
  \caption{The internal structure of an LSTM cell} \label{lstm}
\end{figure}

The standard LSTM model \cite{hochreiter} is composed of one hidden LSTM layer followed by a feed-forward output layer. LSTM differs from other traditional ANNs models in containing memory blocks that replace the summation unit in each cell. The internal structure of an LSTM cell is simply demonstrated in Figure \ref{lstm}.

Each LSTM is a set of arranged cells where the information data stored. In each of these cells, there are some gates built upon a sigmoidal neural network layer allowing each cell to optionally permit the data passing through. Each sigmoidal layer representing the amount of data in each cell within the range zero and one. When the zero value is estimated, this implies that no-information is passing through, whereas if the one value is estimated, this implies that full information is passing through. Each cell has three types of gates:

\begin{itemize}
    \item Forget gate: determines whether the data is removed or retained.  
    \item Memory gate: determines which  data needs to be stored in the cell. 
    \item Output gate: determines which data is useful and can be used for current forecasting. 
\end{itemize}

Recently, the LSTM is used to handle regression via different kinds of energy TSF problems \cite {zhuang, kong, shahzad}. However, there are a few limitations of using LSTM. The first limitation is that the number of memory cells is linked to the size of the recurrent weight matrices. More precisely, an LSTM with $N_h$ memory cells requires a recurrent weight matrix with $O(N_h^ 2 )$ weights, which makes its computation expensive, as the experiments in this paper will show. The second limitation is that LSTM is a poor candidate for learning to represent common data structures like arrays because it lacks a mechanism to index its memory while writing and reading \cite{ivo}. The third limitation is that LSTMs are sensitive to random weight initialization for cells. Recently, the second author of this paper presented a remedy to LSTM limitations in a separate paper \cite{Sagheer2}.   

\subsubsection{Echo state network (ESN)}

The ESN is a type of simplified RNN model uses the idea of reservoir as a medium for information processing to avoid the limitations of RNN, such as expensive computation and slowness. ESN is composed of an input layer, a middle layer, and an output layer. It recovers the RNN limitations by simplifying the learning approach of the network by training only a number of the connected neurons. Once the remaining neuron connections are generated, their weights will not update anymore, where the outputs weights are the only neurons that subject to updating \cite{Tanaka}.

\begin{figure}[h!]
\begin{center}
  \includegraphics[width=0.4\textwidth,height=11pc]{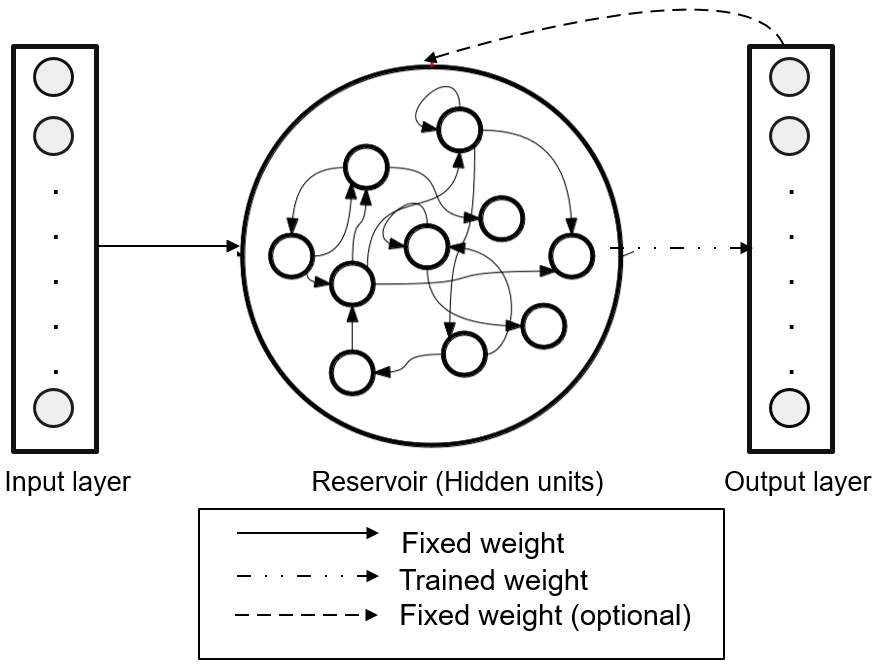}
  \end{center}
  \caption{The ESN architecture} \label{ESN}
\end{figure}

In a comparison with the original RNN, the ESN has a simplified structure and training approach, the advantages that ensure simple and fast training procedure as well as low-cost computation. Compared to other ANN models, the ESN contains a large number of sparsely distributed neuron within the reservoir, as depicted in Figure \ref{ESN}. Though ESN has been applied in many research fields in literature, according to our knowledge, its application to model various energy TSF problems is still limited. Specifically, it is used only to model the TSF problem of wind energy \cite{ESNpaper}.

\subsection{Deep learning methods}
Deep learning is the artificial intelligence branch that causes a significant progress nowadays. Deep learning methods that based on deep architectures of ANNs have repeatedly outperformed the shallow neural network counterparts \cite{DL}. Hence, real applications based on deep learning have been grown rapidly because of the high-performance computing capability of deep learning including the capability for dealing with large size datasets \cite{QZhang}. Therefore, deep learning is more suitable for energy TSF problems as it is easily applicable with large datasets, complex variables, multivariate inputs, along with forecasting multiple time steps \cite{Shi, Tanveer}. 

\begin{figure}[h!]
\begin{center}
  \includegraphics[width=0.45\textwidth,height=14pc]{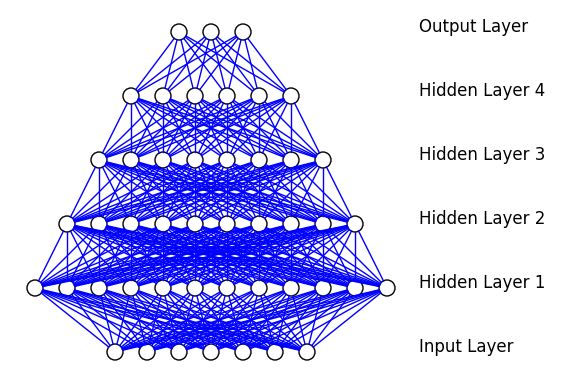}
  \end{center}
  \caption{The deep neural networks architecture} \label{dl}
\end{figure}

The deep NNs have the same structure of the shallow NNs except including more hidden layers, such that each layer processes a portion of the underlying task \cite{Seng}, as shown in Figure \ref{dl}. The role of additional hidden layers is twofold, the first is recombining the learned features from the preceding layers. The second is creating new representations at high levels of abstraction \cite{Seng}. Deep learning algorithms are used efficiently in many energy TSF applications, such as renewable energy \cite{voyant}, solar energy \cite{q}, electricity demand \cite{jaten}, forecasting of load and consumption of electricity \cite{rahman, Shi}. 

The most common way to build a deep network is by stacking more layers one above another with various ways of stacking and learning mechanisms \cite{iclr}. In this paper, we build two deep models, namely, DRNN and DLSTM, in the same way described in our previous works. Therefore, more information about these deep models structure, hyperparameters selection, optimization technique, loss function, and other experimental conditions are given in our previous works \cite{Sagheer2, Sagheer}. These two contributions showed an experimental evidence of the significant benefits of building RNN and LSTM with deep architectures. 

\subsection{Epilog}
We have reached to the end of the first part of this paper, which concerns with the analytical phase. Overall, we reviewed the structures and properties of various conventional, machine learning, and deep learning methods. Specifically, we described briefly the most common methods, in total 15, of each methodology that could be tailored to solve various energy TSF problems, with indications to the existing works. For a qualitative comparison among these methods including the strengthens and limitations of each methods is listed in Table \ref{Overall_Comparison}, at the end of this paper. In the second part of this paper, we will conduct empirical assessment and comparison among all methods that described in the first part, using the same experimental conditions and datasets. This assessment represents a guide to the researchers and practitioners in the energy domain in how should they efficiently select the relevant forecasting model. 

\section{Forecasting Evaluation Metrics}
Evaluating the forecasting model performance is very essential step before selecting the suitable forecast. Of course, the key factor in selecting the appropriate forecasting model is accuracy. The forecast's accuracy is determined by considering how well the chosen forecast model performs on unseen data samples. In a forecasting experiment, the dataset is divided into two disjoint subsections; training and testing. The section of training data is used in fitting the model parameters, whereas the section of testing data is used to calculate the model's prediction. Since the testing data is not utilized during the training model, it should provide a reliable indication about how well the selected model forecasts the unseen data.

For this purpose, many performance measures have introduced to evaluate the model accuracy and calculate the forecast error. The forecast error defines the difference in values between the desired forecast and the actual forecast through the underlying interval of time series. This can be mathematically written as \cite{Srihari}, 
\begin{equation}
e_t= y_t-f_t
\end{equation} 
where \textit{e} is the forecasting error at the time period \textit{t}, \textit{y} is the actual value at the time period \textit{t}, and \textit{f} is the desired forecast value at the same time period \textit{t}. Broadly speaking, forecast errors can be divided into two kinds of errors; residual and prediction errors. The residual errors are those calculated on the training dataset while the prediction errors are those calculated on the testing dataset. In this paper, we are concerned with prediction errors. 

There are two broad kinds of prediction errors \cite{Stanley}.

\begin{itemize}
  \item \textbf{Scale-dependent errors} 
In this kind of error, the forecast errors are on the same scale as the data themselves. The most commonly used measures are the mean absolute error (MAE) and the root MSE (RMSE), where MSE stands for mean square error. They can be represented mathematically as follows,

\begin{eqnarray} \label{MSE}
MSE &=& mean (e_t^2)
\end{eqnarray}
\begin{eqnarray} \label{RMSE}
RMSE &=& \sqrt (MSE)
\end{eqnarray}
  \begin{eqnarray} \label{MAE}
MAE&=& mean (\mid e_t \mid)
\end{eqnarray}

  \item \textbf{Percentage errors} These errors are considered as scale-independent errors, and can be given as  $p_t=e_t/y_t$. The most common measures are the root mean square percentage error (RMSPE), and the mean absolute percentage error (MAPE). They can be represented mathematically as follows,
\begin{eqnarray}
RMSPE&=&\sqrt (mean (p_t^2))*100~(\%)
\end{eqnarray}
\begin{eqnarray} \label{MAPE}
MAPE &=& mean (\mid p_t \mid)*100~(\%)
\end{eqnarray}
\end{itemize}

The forecasting that based on percentage errors has the disadvantage of being undefined if the data contain zero values; i.e. $y_t=0$, in the selected time period. Compared to scale-dependent error, it is widely reported that the percentage error measures are unit-free and more accurate and efficient in tracking the forecasting precision and performance evaluation of the forecasts. The reason behind this is that they have the advantage of being a scale-independent. Accordingly, the measures of percentage error are frequently used in practice to conduct an assessment for different forecasts, particularly when different scaled datasets are used \cite{Hyndman}. For this reason, in the experiment section of this study, we will rely on the values of the two measures MAPE and RMSPE. 

\section{Empirical Assessment}
In this section, we conduct empirical comparison sessions among the the selected models, that described in section 3, in total 15 models. These models are SES, MES, ARIMA, SARIMA, and VAR as representatives for conventional models, kNN, SVR, LSVSR, MLP, ESN, RNNs and LSTM as representatives for machine learning models, and DRNN, and DLSTM as representatives for deep learning models.

The assessments are conducted using three case studies employing three different energy time series datasets, each for a different energy application. Two of these datasets are including univariate time series observations, whereas the third dataset includes a multivariate time series observations. In each experiment, we will describe the employed dataset and then show its related forecasting results.

It is not logic; and maybe unfair, to assess and compare all models in the three methodologies on the same base in spite of differences in nature and architecture of each model. Therefore to make a convincing assessment, in each case study first we will compare the performance of conventional methods against the standard machine learning models. Then, we will compare the performance of standard machine learning (or shallow neural networks) models against deep learning (or deep neural networks) models.

\subsection{Hardware and software platforms}

All experiments in this paper are implemented on an HP workstation-PC equipped with Ubuntu 16.04 operating system. The computational time of each algorithm was estimated using a system with Intel Core™ i7-6700 CPU @ 3.40GHz, 8.00 GB RAM, x64 based processor under python 2.7 software environment. For ANNs methods, the Keras library was used with an open-source TensorFlow \cite{Tensor} library as back-end. 

\subsection{Case Study-I: Univariate Time Series}
This case study concerns an electrical energy consumption problem.
\subsubsection{Description of dataset}
\begin{figure}[h!]
\begin{center}
  \includegraphics[width=0.5\textwidth,height=12pc]{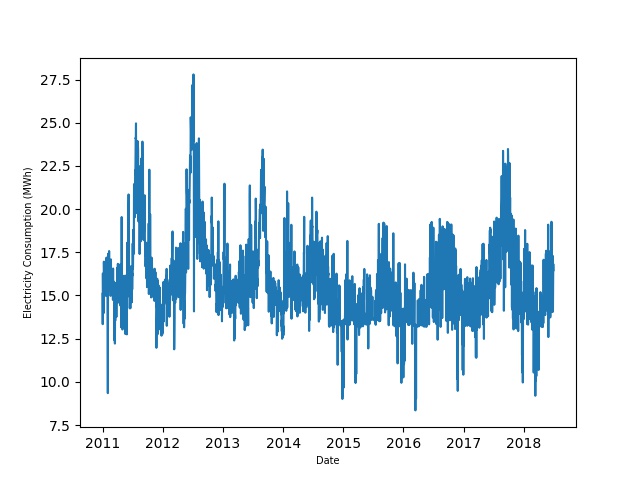}
  \end{center}
  \caption{CBU Monore intake tower power-daily consumption} 
  \label{mono-day}
\end{figure}

\begin{figure}[h!]
\begin{center}
  \includegraphics[width=0.5\textwidth,height=12pc]{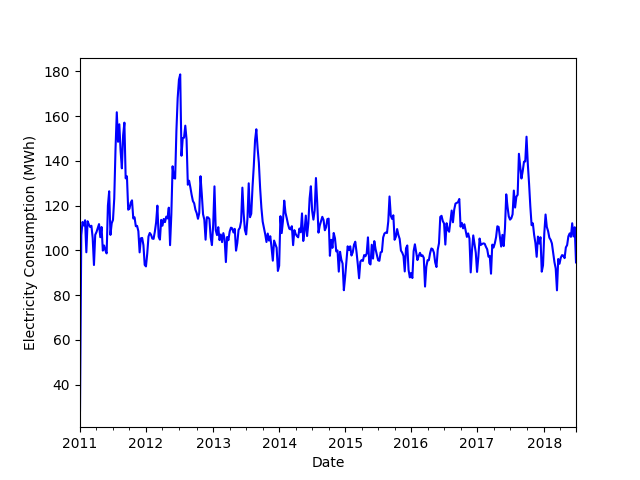}
  \end{center}
  \caption{CBU Monore intake tower power-weekly consumption} 
    \label{mono-week}
\end{figure}

The data samples of this dataset were collected from the City of Bloomington Utilities (CBU)-intake tower. This facility pumps water from lake Monore to Monore water treatment plants, southern U.S state, Indiana. The daily energy consumption used at this facility is measured in (MWh) from January 2011 to June 2018. The data are available for public use on the website \cite{Bloom}. The dataset has 2,738 indexes splitting into a training dataset (67\%) and testing dataset (33\%). The weekly power consumption of the data is shown in Figure \ref{mono-day} and Figure \ref{mono-week} for daily and weekly consumption, respectively.

\subsubsection{Experimental results}
Table \ref{SN_mon} shows the performance of the conventional methods and the machine learning methods, along with the best parameter values for each method. It is clear that the machine learning models show a growing improvement over the conventional methods on the scope of the three performance metrics. If we select one metric; for example MAPE, we will notice that SES showed 8.75\%, ARIMA showed 8.53\%, LSSVR showed 5.44\%, MLP showed 6.17\%, ESN showed 7.9\%, RNN showed 5.32\%, and LSTM showed 5.19\%. The same performance can be observed in the metrics RMSE and RMSPE. Not only outperformed the conventional counterparts, but the LSTM also improved the forecasting error compared to machine learning counterparts; ESN, RNN and MLP, even if the difference with RNN is slightly scanty. The same attitude can be observed in the results of the RMSPE measure. On the level of RMSE measure, the same attitude is shown where both ARIMA and SES showed about 1.7\%, MLP showed 1.31\%, ESN showed 1.54, \%whereas both RNN and LSTM showed the same error rate about 1.06\%.

\begin{table*}[ht]
\centering 
\caption{Comparison among conventional and machine learning models for case study-I. NoL: No of layers; NoN: No of neurons}
\label{SN_mon}

\begin{tabular}{l l l l l l }
\cline{1-6}
  & & &  \multicolumn{3}{l }{~~~~~ Performance Measure} \\\cline{4-6}
 {Method} & {NoL} &  {NoN} & {RMSE} & {RMSPE} &{MAPE}\\\cline{1-6}
{SES  }($\alpha$=0.4) &-&- &1.59 & 10.79 &8.15\\ \cline{1-6}
{ARIMA}(p,d,q=1,1,0)&-&-&1.69 & 11.42 &8.53\\ \cline{1-6}
{SARIMA}&-&- &1.43&9.81&7.55\\ \cline{1-6}
{kNN} (n=3)&-&- &1.67&10.28&7.9 \\ \cline{1-6}
{SVR}  &-&-&1.62&9.81&6.05 \\ \cline{1-6}
{LSSVR}  &-&-&1.54&9.07&5.44 \\ \cline{1-6}
{MLP}&4 &12& 1.31&8.43 &6.17\\ \cline{1-6}
{ESN}&- &50& 1.54&10.39 &7.90\\ \cline{1-6}
{RNN}  &1 &2& 1.07 &7.02 &5.32\\ \cline{1-6}
{LSTM}&1&3&\textbf{1.06} &\textbf{6.95} &\textbf{5.19}\\

\hline
\end{tabular}
\end{table*}

\begin{table}[h!]
\centering 
\caption{Comparison between RNN and DRNN for case study-I}
\label{drnn_mon}
\begin{tabular}{l l l l l l }
\cline{1-6}
  & & &  \multicolumn{3}{l }{~~~~~ Performance Measure} \\\cline{4-6}
{Method} & {NoL} &  {NoL} & {RMSE} & {RMSPE} &{MAPE}
\\ \cline{1-6}
 \multirow{1}{*}{RNN} &1 &2& 1.07 &7.02 &5.32\\

\cline{1-6}

 \multirow{1}{*}{DRNN} 
&3 &6& 0.57 &\textbf{3.47} &\textbf {2.71}\\ \cline{1-6}

\hline
\end{tabular}
\end{table}

Since the RNN and LSTM demonstrated the smallest forecasting errors over other ANN counterparts, we examined their performance when deep architectures are adopted. Table \ref{drnn_mon} and Table \ref{dlstm_mon} display the comparison between shallow architectures and deep architectures for both RNN and LSTM, respectively. It is clear that the deep models improved the performance of standard or shallow models. If we select the RMSPE metric, the RNN showed around 7\% whereas DRNN showed around 3.5\%, and the LSTM showed around 6.9\% whereas DSLTM showed around 3.3\%. 

\begin{figure}[h!]
    \centering
    \includegraphics[width=0.5\textwidth,height=11pc]{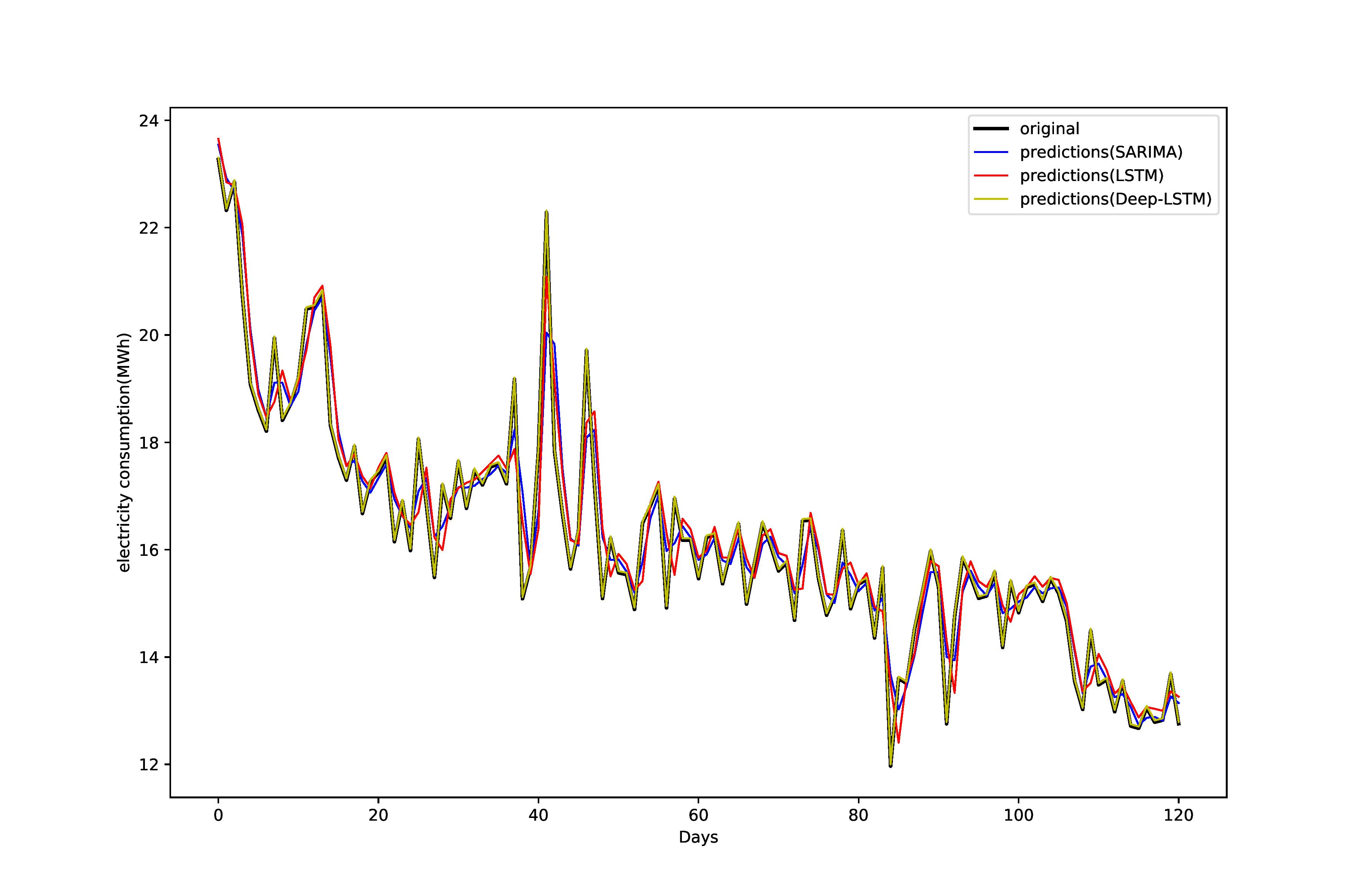}
    \caption{The consumption prediction during the last four months of the first year for the methods ARIMA, LSTM, and DLSTM}
    \label{comp1}
\end{figure}

The improvement in forecasting errors brought by deep models is clearly high compared to forecasting errors brought by shallow models. Indeed, this assessment confirms the superiority of deep learning models over conventional and machine learning models. For visual comparisons, Figure \ref{comp1} illustrates the consumption prediction of the last four months of the first year using the methods SARIMA, LSTM, and DLSTM. We selected these three methods in this specific period in order to have a clear visual illustration. 

\begin{table}[h!]
\centering 
\caption{Comparison between LSTM and DLSTM for case study-I}
\label{dlstm_mon}
\begin{tabular}{l l l l l l }
\cline{1-6}
  & & &  \multicolumn{3}{l }{~~~~~ Performance Measure} \\\cline{4-6}
 {Method} & {NoL} &  {NoN} & {RMSE} & {RMSPE} &{MAPE}\\ \cline{1-6}
\multirow{1}{*}{LSTM}&1&3&1.06 &6.95 &5.19\\\cline{1-6}
\multirow{1}{*}{DLSTM} & 4 & 12&\textbf{0.52}&\textbf{3.28}&\textbf{2.45}\\\cline{1-6}
\hline
\end{tabular}
\end{table}

\subsection{Case Study-II: Univariate Time Series}
This case concerns a natural gas consumption problem.
\subsubsection{Description of dataset}

\begin{figure}
\centering
    \includegraphics[width=0.5\textwidth,height=11pc]{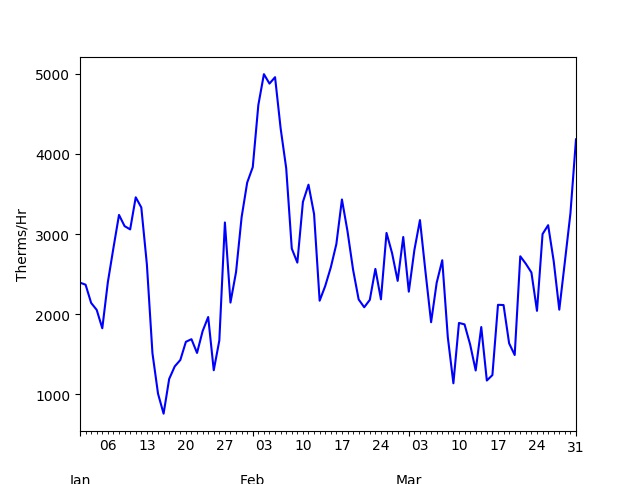}
    \caption{The daily natural gas consumption at BNC}
    \label{monoweek}
\end{figure}

The data samples of this case study were collected for 5-minute intervals for three months from January to March 2014. It was the natural gas consumption of a building number 74 located in the Lawrence Berkeley National Lab campus (BNC). Lawrence Berkeley National Laboratory (Berkeley Lab) is a department of Energy (DOE) office of Science lab managed by the University of California. Since natural gas meters measure volume and net energy content, a therm factor is used by natural gas companies to convert the volume of gas used to its heat equivalent, and thus calculate the actual energy use. So, BNC uses natural gas with consumption measured in (Therms/hr). The data have 25,908 indexes splitting into a training dataset (67 \%) and testing dataset (33\%). The daily natural gas consumption at BNC is shown in Figure \ref{monoweek}.

\subsubsection{Experimental results}
We will follow the same scenario adopted in the previous case study. Table \ref{SN_bc} displays the performance of the conventional methods and the machine learning methods, with the best parameter values for each model. Newly, the machine learning models show a growing improvement over the conventional methods in the scope of the three performance metrics. Let us select this time the RMSPE measure, we will notice that ARIMA showed 18.6\% whereas SES showed about 18.1\%. However MLP and ESN are machine learning models, they showed errors with 17.61\% and 16.98\%, respectively, which are very close to the performance of conventional models. In contrast, RNN showed 2.9\% and LSTM showed about 3.1\% in a significant improvement compared to other contenders. The same attitude can be observed in the results of the MAPE measure. On the level of RMSE measure, the same attitude is shown where ARIMA, SES, ESN, and MLP showed about 0.7\%, whereas a big improvement bought by both RNN and LSTM which showed the same error rate about 0.1\%.

\begin{table*}[ht]
\centering 
\caption{Comparison among conventional and machine learning models on dataset-II. NoL: No of layers; NoN: No of neurons}
\label{SN_bc}
\begin{tabular}{l l l l l l }
\cline{1-6}
  & & &  \multicolumn{3}{l }{~~~~~ Performance Measure} \\
\cline{4-6}
{Method}  & {NoL} &  {NoN} & {RMSE} & {RMSPE} &{MAPE}\\ \cline{1-6}
{ARIMA }(p,d,q=1,1,0)  &- &- & 0.74& 18.59 &10.35\\ \cline{1-6}
{SES  }($\alpha$=0.6)&- &- &0.76& 18.07 &10.07\\ \cline{1-6}
{SARIMA}  &- &- &0.58&16.4&8.91 \\ \cline{1-6}
{kNN }(k=3) &- &- &0.8&16.2&11.08  \\ \cline{1-6}
{SVR} &- &- &0.47&6.82&4.83   \\ \cline{1-6}
{LSSVR}  &- &- &0.31&6.2&3.90\\ \cline{1-6}
{MLP} &4 &12& 0.75&  17.61&11.09\\ \cline{1-6}
{ESN}&- &100& 0.69&16.98&9.78\\ \cline{1-6}
{RNN}&1 &3& \textbf{0.09}&\textbf{2.91} &1.68\\ \cline{1-6}
{LSTM} &1 &3& \textbf{0.09}&3.06 &\textbf{1.78}\\\hline
\end{tabular}
\end{table*}

For the deep learning models, Table \ref{drnn_bc} and Table \ref{dlstm_bc} display the comparison between shallow architectures and deep architectures for each model separately. Again, it is clear that the deep architectures improve the performance of shallow architectures. If we select the RMSPE measure, the RNN showed 2.9\% whereas DRNN showed about 2.1\%. Also, the LSTM showed about 3.1\% whereas DSLTM showed about 1.9\%. Certainly, the improvement in forecasting brought by the deep models is high compared to shallow ones, which confirms their superiority over conventional and standard machine learning models. 
 
\begin{table}[h!]
\centering 
\caption{Comparison between RNN and DRNN for case study-II}
\label{drnn_bc}
\begin{tabular}{l l l l l l }
\cline{1-6}
  & & &  \multicolumn{3}{l }{~~~~~ Performance Measure} \\\cline{4-6}
 {Method} & {NoL} &  {NoN} & {RMSE} & {RMSPE} &{MAPE}\\ \cline{1-6}
 \multirow{1}{*}{RNN} &1 &3& 0.09&2.91 &1.68\\\cline{1-6}
 \multirow{1}{*}{DRNN} &3 &6& \textbf{0.07}&\textbf{2.13}&\textbf{1.44}\\ \cline{1-6}\hline
\end{tabular}
\end{table}

\begin{table}[h!]
\centering 
\caption{Comparison between LSTM and DLSTM for case study-II}
\label{dlstm_bc}
\begin{tabular}{l l l l l l }
\cline{1-6}
  & & &  \multicolumn{3}{l }{~~~~~ Performance Measure} \\\cline{4-6}
 {Method} & {NoL} &  {NoN} & {RMSE} & {RMSPE} &{MAPE}\\ \cline{1-6}
\multirow{1}{*}{LSTM} &1 &3& 0.09&3.06 &1.78\\\cline{1-6}
\multirow{1}{*}{DLSTM} & 4 & 12 &\textbf{0.06}& \textbf{1.95} &\textbf{0.92} \\\cline{1-6}\hline
\end{tabular}
\end{table}

\begin{figure}
    \centering
    \includegraphics[width=0.5\textwidth,height=10pc]{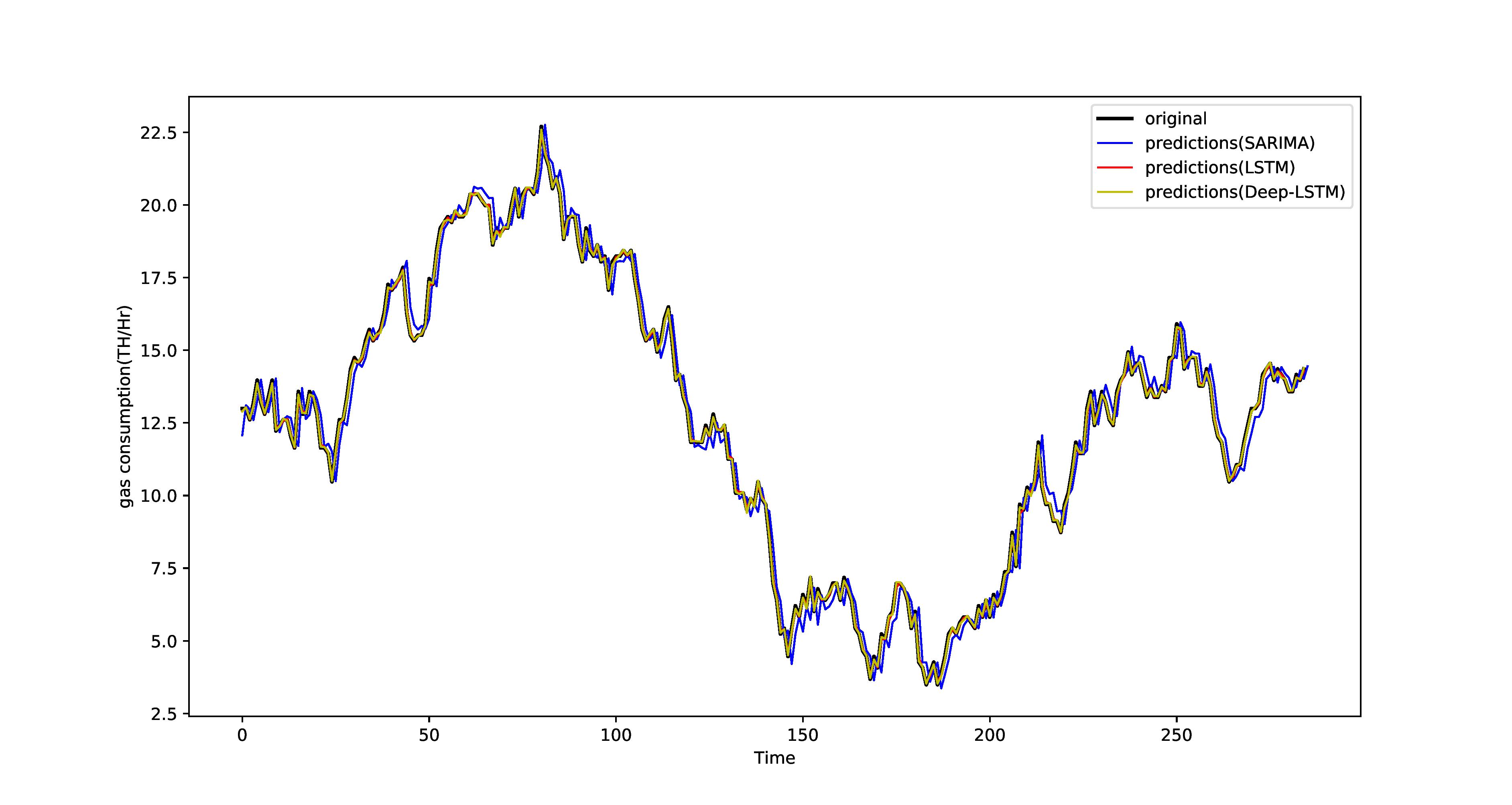}
    \caption{The consumption prediction of the last day in the first month for the methods SARIMA, LSTM, and DLSTM}
    \label{lastday}
\end{figure}

\subsection{Case Study-III: Multivariate Time Series}
This case study concerns the electric power consumption problem of an individual household. The sector of individual house consumption is one of the largest consumers of electric energy. The rational consumption of electricity at home becomes of a great importance \cite{EMS}. Accordingly, this case study is considered as a large scale TSF problem.  

\subsubsection{Description of dataset}
The household power consumption dataset is a multivariate time series dataset that describes the electricity consumption for a single household over four years, exactly 47 months. The dataset instances were collected every minute, which yields a total number of instances equal to 2075259 instance \cite{UCI}. Beside the time and date, the dataset identifies the following seven variables of interest: global-active-power (the total active power consumed by the household), global-reactive-power (the total reactive power consumed by the household), voltage (Average voltage), global-intensity (average current intensity), sub-metering-1 (active energy for kitchen), sub-metering-2 (active energy for laundry), and sub-metering-3 (active energy for climate control systems). The active energy variable is the real power consumed by the household, whereas the reactive energy variable is the unused power in the electric lines.

\begin{figure}
    \centering
    \includegraphics[width=0.47\textwidth,height=13pc]{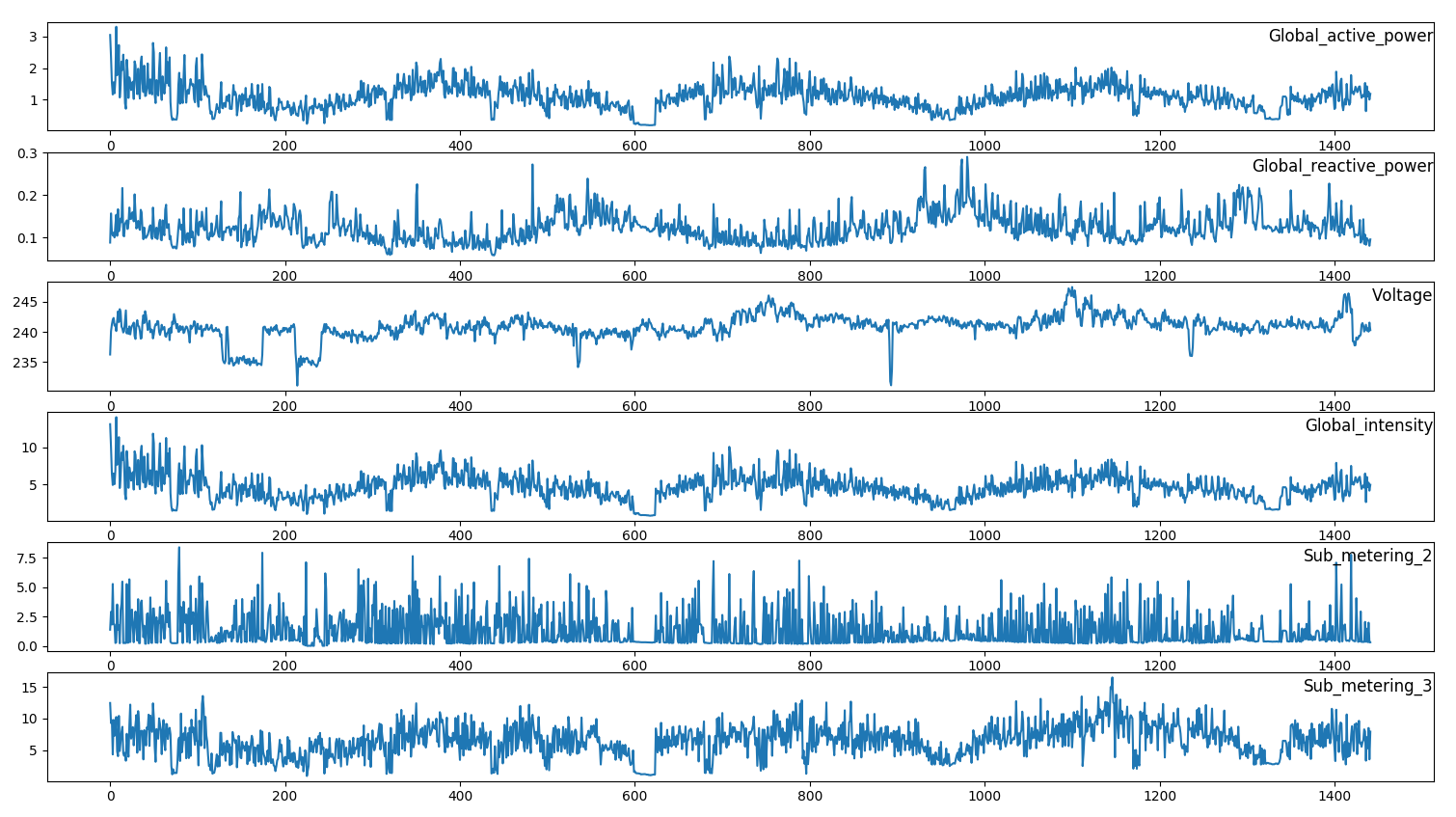}
    \caption{Visualization of the decomposed variables of the household power consumption (multivariate) dataset}
    \label{7-variables}
\end{figure}

The best way to understand these two core variables is graphical visualization. By this way, we are able to detect if the data contains a significant trend or some consistent or irregular attributes as well as understanding seasonality. For this purpose, we create a separate plot for each of the seven variables, as shown in Figure \ref{7-variables}. If we look closely to the first panel of global-active-power, it is clear that it is strongly annual seasonal dependent. In addition, some trend can be noticed there as well as some random aspects influences the data distribution. Specifically, we can easily notice a downward trend over the summer months (middle of the year) and more consumption in the winter months (at the edges of the plots). Same phenomena we may notice to some extent in the second panel of global-reactive-power.

\subsubsection{Experimental results}
All the aforementioned properties of the dataset attribute the nonlinearity and complexity of the current case study. Due to voluminous size and nonlinearity of the employed dataset, this dataset is very ideal for evaluating the deep learning models against conventional methods. As we explained in section 3.1. since the current problem is a multivariate regression problem, we will use VAR instead of ARIMA and SARIMA, and MES instead of ES and SES. Also, due to technical limitations of kNN and SVR to model multivariate regression, we excluded them in this experiment.

Table \ref{MTV-results} shows the performance of the conventional methods and the machine learning models. After the first look to the table, we will observe that the error values are big, particularly the values of RMSPE and MAPE metrics, as a reflection to the bigger values of the data samples \cite{UCI}. Regardless of the values voluminous, it is clear that the machine learning models outperformed the conventional methods on the context of the three performance metrics. If we select the MAPE metric, we will notice that MES showed 103.66, whereas VAR showed 89.64. In contrast, MLP showed 67.62, RNN showed 45.91, and LSTM showed 46.50. The same performance can be observed in the other metrics RMSE and RMSPE. Overall, we can easily notice that the RNN improved the forecasting error compared to other machine learning models; LSTM and MLP, and compared to conventional methods. 

\begin{table}[ht]
\centering 
\caption{Comparison among conventional and machine learning models on case study-III. NoL: No of layers; NoN: No of neurons}
\label{MTV-results}
\begin{tabular}{l l l l l l }
\cline{1-6}
  & & &  \multicolumn{3}{l }{~~~~~ Performance Measure} \\\cline{4-6}
{Method}  & {NoL} &  {NoN} & {RMSE} & {RMSPE} &{MAPE}\\ \cline{1-6}
\multirow{1}{*}{MES}  &- &-& 1.14 & 119.27& 103.66\\ \cline{1-6}
\multirow{1}{*}{VAR}  &- &-&0.87 &106.78&89.64\\ \cline{1-6}
\multirow{1}{*}{MLP}  &3 &100& 0.69&  79.32&67.62\\ \cline{1-6}
\multirow{1}{*}{RNN}  &1 &50& \textbf{0.56}&\textbf{64.22} &\textbf{45.91}\\ \cline{1-6}
\multirow{1}{*}{LSTM}  &1 &50& \textbf{0.56}& 66.08 &46.50\\\hline
\end{tabular}
\end{table}

\begin{figure}
    \centering
    \includegraphics[width=0.48\textwidth,height=13pc]{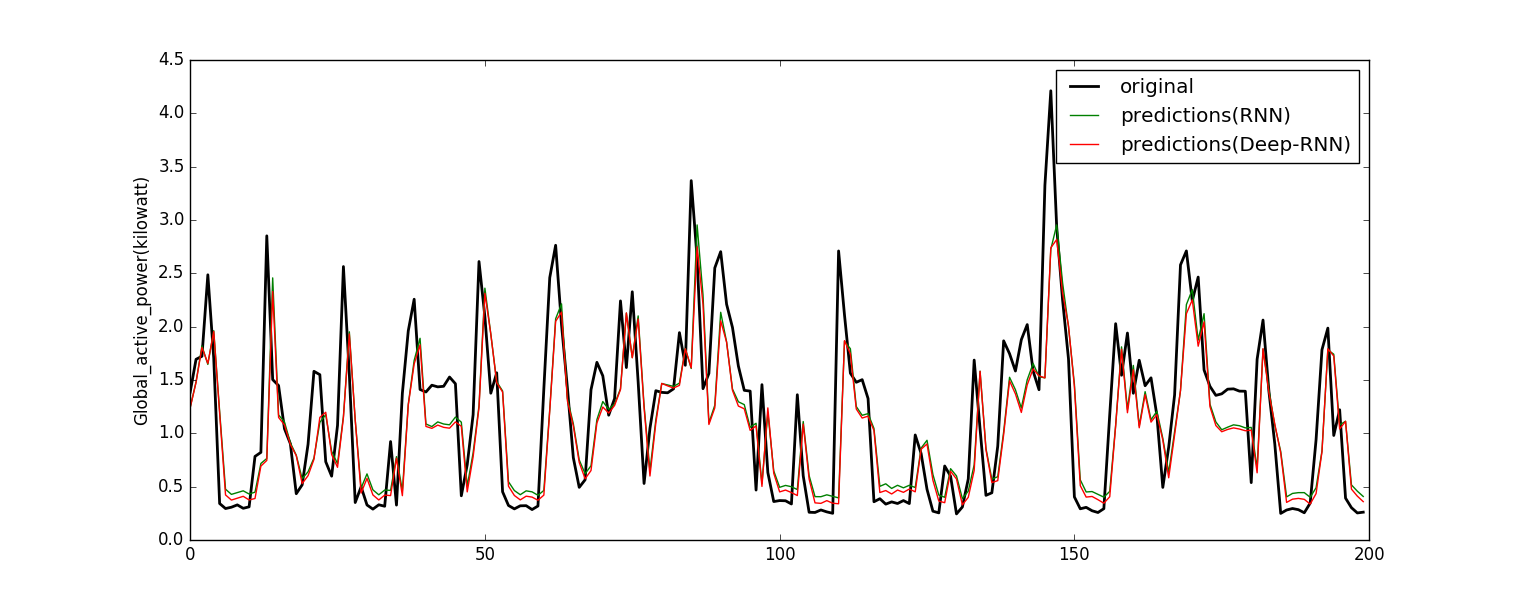}
    \caption{Global active power prediction for RNN and DRNN}
    \label{rnn_drnn_multi}
\end{figure}

To examine the impact of deep architectures, we applied the DRNN and DLSTM in this case study as well. Table \ref{drnn_III} and Table \ref{dlstm_III} display the comparison between shallow and deep architectures for each RNN and LSTM, respectively. It is clear that the deep models improve the performance of shallow models. On the scale of the RMSPE metric, the RNN showed an error rate as 64.22 whereas DRNN showed 62.57. Similarly, the LSTM showed an error rate 66.08 whereas DSLTM showed 64.71. The same attitude can be noticed on the scale of the MAPE metric, however, all models showed the error rate 0.56 on the scale of RMSE metric. Figure \ref{rnn_drnn_multi} shows a visualization of the performance of the shallow RNN against the performance of DRNN. Doubtless, including many inputs, or variables, considerably increases the computation complexity of a model. This case study confirms the superiority of deep learning models compared to other contenders in terms of forecasting accuracy.

\begin{table}[h!]
\centering 
\caption{Comparison between RNN and DRNN for case study-III}
\label{drnn_III}
\begin{tabular}{l l l l l l }
\cline{1-6}
  & & &  \multicolumn{3}{l }{~~~~~ Performance Measure} \\\cline{4-6}
 {Method} & {NoL} &  {NoN} & {RMSE} & {RMSPE} &{MAPE}\\ \cline{1-6}
 \multirow{1}{*}{RNN} &1 &50& 0.56&64.22 &45.91\\\cline{1-6}
 \multirow{1}{*}{DRNN} &3 &100&0.56&\textbf{62.57}&\textbf{43.88}\\ \cline{1-6}\hline
\end{tabular}
\end{table}

\begin{table}[h!]
\centering 
\caption{Comparison between LSTM and DLSTM for case study-III}
\label{dlstm_III}
\begin{tabular}{l l l l l l }
\cline{1-6}
  & & &  \multicolumn{3}{l }{~~~~~ Performance Measure} \\\cline{4-6}
 {Method} & {NoL} &  {NoN} & {RMSE} & {RMSPE} &{MAPE}\\ \cline{1-6}
\multirow{1}{*}{LSTM} &1 &50& 0.56& 66.08 &46.50\\\cline{1-6}
\multirow{1}{*}{DLSTM} & 3 & 100 &0.56& \textbf{64.71} &\textbf{44.65} \\\cline{1-6}\hline
\end{tabular}
\end{table}

\section{Results Discussion and Analysis}
Based on the empirical assessments described in the previous section, it is clear that the deep learning models, represented in DRNN and DLSTM, outperformed both the machine learning models and the conventional models. In this section, we analyze the performance of all models from different aspects.  

\subsection{Analysis in terms of accuracy and performance}

\begin{figure}[h!]
    \centering
    \includegraphics[width=0.5\textwidth,height=15pc]{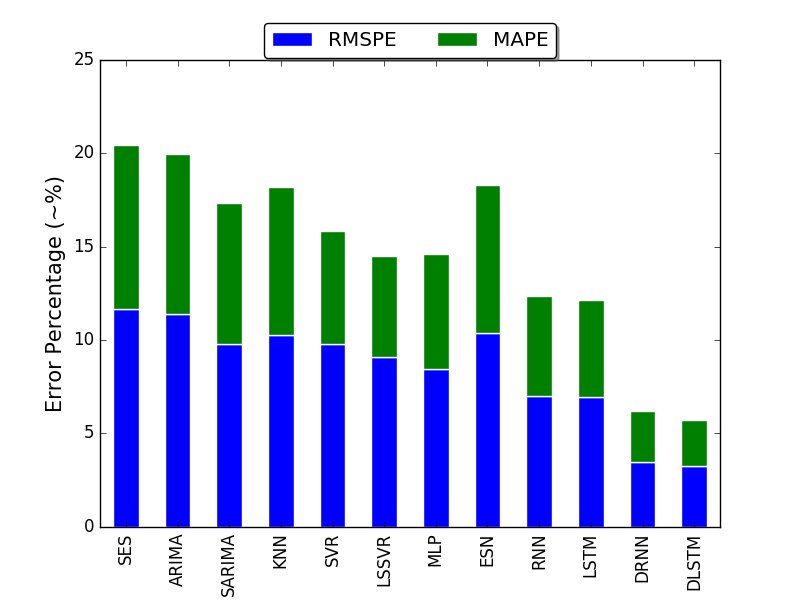}
    \caption{Percentage errors RMSPE and MAPE for Case Study-I }
    \label{err1}
\end{figure}

\begin{figure}
    \centering
    \includegraphics[width=0.5\textwidth,height=15pc]{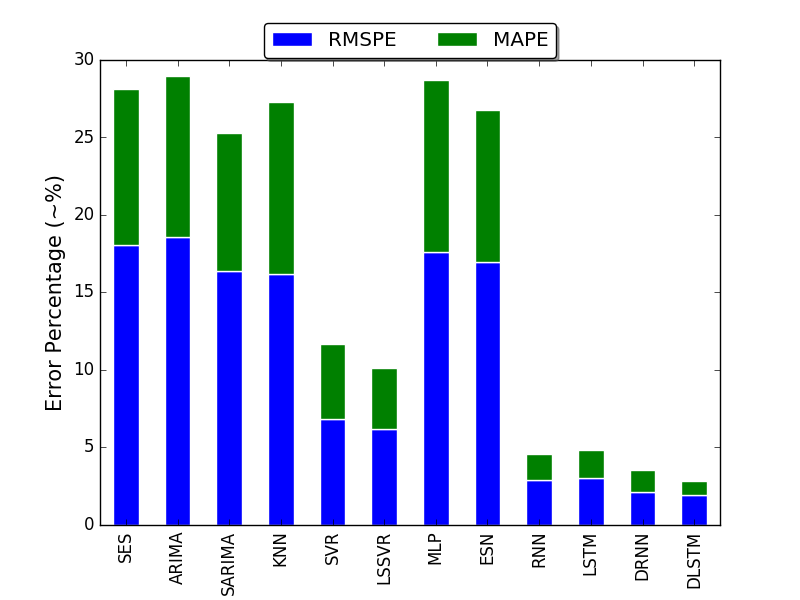}
    \caption{ Percentage errors RMSPE and MAPE for Case Study-II  }
    \label{err2}
\end{figure}

\begin{figure}
    \centering
    \includegraphics[width=0.5\textwidth,height=15pc]{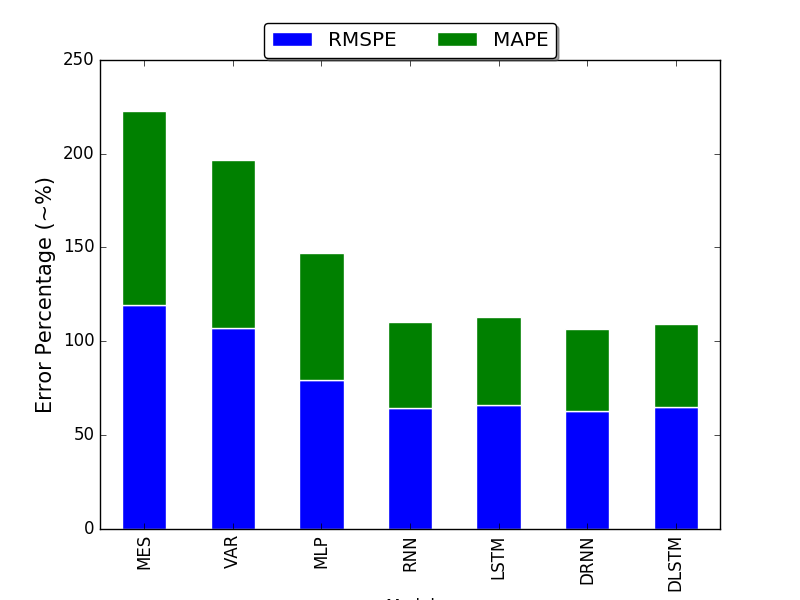}
    \caption{ Percentage errors RMSPE and MAPE for Case Study-III  }
    \label{err3}
\end{figure}

No doubt that, the key factor in selecting the appropriate forecasting model is the accuracy and overall performance. Beside the presented tables, Figures \ref{err1}-\ref{err3} demonstrate visual representation for the percentage errors, MAPE and RMSPE, of all models. It is clear that the errors shown by DLSTM and DRNN are lower than those shown by all other contenders. This performance is steady through the three case studies and did not change according to the employed dataset.

The figures point out that the performance of conventional methods (ARIMA, SES, SARIMA, VAR, MES), across the three case studies, is poor. This performance asserts that the conventional methods could not effectively extract the features from a time series dataset. The same poor performance, but with a little improvement, is shown by MLP, kNN, SVR, and LSVSR. In contrast, consistent improvement in performance is shown by the shallow neural network models, while a significant improvement is brought by their deep counterparts. The reason for this improvement brought by deep models over the shallow models is that, the deep NN learns the nonlinear combinations, or correlations, of the features in higher layers of the network. However, the hidden neurons in shallow NN models learn the nonlinear combinations of the inputs as the features. Therefore, the deep architecture will learn the hierarchical features, for example in the load demand problem, at different layers smoother than shallow architecture  \cite {Shi}. 

Specifically, the significant improvement brought by DRNN and DLSTM is attributed to the properties of these models. They exploit the benefits of memory loops that integrated in a deep architecture, enabling each network to learn from a long horizon dataset \cite{Herman, iclr, DL}. In addition, both models are relying on the dynamic (or active) learning mode, in which the model integrates the historical observations with the recent ones to efficiently predict the future \cite {jaten}. This property is very useful particularly when process an energy demand problem with long horizon. On the contrary side, the conventional methods and other machine learning methods are static learning based approaches. Actually, they do not rely on the explicit relationship among the historical data and future data. Rather they just learn from the available historical observations for prediction \cite{jaten}. Such a static learning mode faces troubles when a long-term forecasting is needed. Certainly, the longer the forecasting horizon, the greater is the possibility of a forecasting error.

\subsection{Analysis in terms of time series data}
The presented results indicate the relation between the forecasting error and the size of the dataset. Precisely, given that the size of the first dataset is small (2738 samples), the conventional methods showed the level of 11.5\% on the RMSPE measure and 8.5\% on the MAPE measure. Once the size of the dataset is getting larger, such as the second dataset that has 25908 instances, the error rates jumped to approach the level of 18\% on RMSPE measure and 10\% on the MAPE measure. When the size of the dataset became very voluminous with multiple variables included; such as the third dataset that has 2075259 instances, the difference in performance became clear with a superiority for deep learning models. This performance gives a credit for deep models in treating a broad spectrum of prediction applications, particularly if the observations correlate many variables \cite{QZhang}.    

Looking inside each dataset we will notice that the observations of all datasets were recorded in short intervals, every one and five minutes for dataset I and III, respectively, and every one hour for dataset II. Keeping in mind their poor performance, the conventional methods are not fit for such short interval observations. Actually, this conclusion is consistent with the result reported in \cite {NanWei} who stated that conventional methods are preferred for the yearly energy consumption. In line with the latter conclusion and keeping in mind their well performance, the deep learning models are more robust and stable in all data intervals and forecasting horizons. They providing the optimum forecasting for either short-term or long-term energy consumption. Surprisingly, this contradicts with what Wei. et al have reported that deep learning models need further validation compared to the conventional methods \cite{NanWei}. 

In terms of data structure, the machine learning, including deep learning, models have a strong capability to handle nonlinear and heterogeneous time series data, either short-term or long-term forecasting problems. In the contrary, the conventional methods suffer when handling nonlinearity and nonstationerity in time series data. Nevertheless, it is worth to highlight that the conventional methods, in general, do not need many of historical data to function, contrary to the learning-based models, which require plentiful data to learn.

\subsection{Analysis in terms of computational complexity}

Toward a rational assessment, we calculated the computational complexity (CC) of each method in order to delimit the time taken to fit a model and use it for prediction. In the same way described in \cite{mar}, the formula of CC for a specific method to accomplish a specific task can be represented as shown in Eq. (\ref{cc}). As the denominator in Eq. (\ref{cc}) is unified for all computations, we can rely only on the numerator value and, eventually, we shall have a value indicates the proportional time taken by a model to achieve the prediction. 
\begin{equation}\label{cc}
CC=\frac{Model~Computational~Time}{Naive~Computational~Time}
\end{equation}

\begin{table}[h!]

    \centering
    \caption{Model computational time in seconds}
    \label{CC}
    \begin{tabular}{l l l}
    \hline
        Method &~~~~~~~~~ Fitting Time &~~~~~~~~~ Prediction Time \\\hline
        ARIMA  &~~~~~~~~~ 0.02&~~~~~~~~~ 0.0007\\ \hline
        SES &~~~~~~~~~ 0.005  &~~~~~~~~~ 0.002\\ \hline
        SARIMA &~~~~~~~~~ 0.016  &~~~~~~~~~ 0.07 \\ \hline
        SVR &~~~~~~~~~ 0.17  &~~~~~~~~~ 0.03 \\ \hline
        LSSVR &~~~~~~~~~ 0.09  &~~~~~~~~~ 0.002 \\ \hline
        kNN &~~~~~~~~~ 0.002  &~~~~~~~~~ 0.001 \\ \hline
        ESN &~~~~~~~~~ 0.056  &~~~~~~~~~ 0.028\\ \hline
        MLP &~~~~~~~~~ 2616  &~~~~~~~~~ 0.14  \\ \hline
       RNN  &~~~~~~~~~ 3151  &~~~~~~~~~ 0.17 \\ \hline
       LSTM &~~~~~~~~~ 4318 &~~~~~~~~~  0.33\\ \hline
      DRNN  &~~~~~~~~~ 4542  &~~~~~~~~~ 0.33 \\ \hline
     DLSTM  &~~~~~~~~~ 10208 &~~~~~~~~~ 0.59 \\ \hline
    \end{tabular}
    
\end{table}
To avoid redundancy, we will settle for calculating the CC required by each model using only the samples of case study-I, as shown in Table \ref {CC}. The fitting time is the time required to train the model, whereas the prediction time is the time required to test or validate the model. It is clear that the computation requirements for conventional methods are very cheap compared to other contenders, particularly deep models. For example, the prediction time of ARIMA and SES are about 0.001 and 0.002 seconds, respectively, whereas DRNN and DLSTM need 0.33 and 0.6 seconds to predict, respectively. No doubt, this is a big privilege of conventional methods contrary to other models. 

On the same time, the fastest machine learning model is kNN that requires 0.001 seconds to predict. The fastest among ANN models is ESN, which requires about 0.03 seconds to converge. Indeed, the ESN time requirement is reasonable since it is already developed as a faster version of RNN. In the meantime, it is totally credible that the deep learning models require longer time to function in terms of increasing number of hidden layers and neurons. Another negative aspect of increasing the number of layers in deep models may reflect the occurrence of overfitting, especially in lack of data diversity \cite{Shi}. Nevertheless, using convolution optimization techniques may solve the overfitting problem by optimally select the proper parameters of the deep model according to the problem at hand \citep{Yang}. In addition, with the emerging progress in hardware and learning algorithms, there are many remedies for the expensive computation cost of deep learning models \cite {Chandra, Cong}.  

\subsection{Epilog}
Overall, the main finding of this paper is the suitability of deep learning models to be applied in various energy problems. The given three case studies confirm the stability, well-performance, and robustness of deep learning models; DRNN and DLSTM. Accordingly, there are no reasons for some recent articles to assume that deep learning is not stable as it does not test yet \cite{NanWei}. We are very certain that the adoption of deep learning methods, in the energy domain, is right now at the same maturity level as conventional methods, which are the dominant methodologies for decades.

\section{Challenges for Future Research}
Though the comprehensive analytical and empirical review presented in this paper, there are a number of challenges represent open-ended questions.   

\begin{enumerate}
\item The probabilistic prediction of various energy problems has not attracted enough attention, however, it is a vital research domain that should be addressed. In energy management systems, the probabilistic predictions can provide a range of energy changes that may help to quantify the uncertainty involved \cite{EMS}. 

\item Most of the research in the energy TSF domain consider only the historical data and neglect other influencing or exogenous factors. For example, in energy consumption problems, there are many factors that influence the rates of consumption in a building or a city, including weather conditions, human occupancy, and indoor conditions. These influencing factors and many others need to be taken into account in order to have efficient forecasting results.

\item In the scope of their robust mathematical and statistical foundations, conventional methods able to show a good performance in terms of representing the relationship between historical data and the influencing factors. In contrast and in the scope that ANNs are black-box approaches, they might not have the same ability.

\item Handling a collaborative training on multiple energy prediction tasks is very essential in the future of energy research.
\end{enumerate}

All these challenges represent future research directions should be further investigated by researchers. However, there are a number of extensions to the work presented in this paper could be addressed as well. For example, the forecasting problem of multiple variables needs deep investigation, particularly in case of missing and distorted sequential observations. In addition, various optimization techniques can be adopted in order to improve the prediction accuracy of deep learning models. Moreover, developing an intelligent prediction platform, with a comprehensive auto-analysis and visualizations of the useful insights, represents a great enhancement toward intelligent energy management systems \cite{EMS}. 

\section{Conclusions and Recommendations}
Time series forecasting problem has a remarkable importance in various practical and industrial applications nowadays. Various methodologies have widely employed to solve different time series forecasting problems. This paper presents a qualitative analytical review along with an quantitative empirical assessment for the conventional, machine learning, and deep learning methodologies that applicable in the energy TSF domain. In the analytical review part, we could not pretend to have reviewed all existing models in this wide domain, however, we selected the most remarkable models, in total 14, along with the advantages and disadvantages of each model, as shown in Table \ref{Overall_Comparison}, at the end of this paper. In the empirical assessment part, we conducted several experiments using the selected models. Three real world datasets in the energy domain have used, two datasets include univariate observations and one dataset includes multivariate observations. Except the high computation requirements, the analytical and empirical assessments indicate the superiority of the deep learning models in terms of accuracy and forecasting horizons examined,
compared to the other contenders. Ideally, the outcome of this study should motivate the AI and machine learning community for further development in this trend of research. 

Based on the aforementioned qualitative and quantitative assessments, we can suggest the following recommendations:

\begin{enumerate}
\item The deep learning models DRNN and DLSTM are robust and stable enough to be applied in all forecasting horizons of energy applications. In addition, they support the dynamic learning mode, in which the model integrates the historical observations with the recent ones to predict the future. Unlike other machine learning methods and conventional methods that relay directly on the historical observations only, which is known as static learning.

\item It is possible that hybrid methods, which combine one or more conventional and machine learning methods, outperform a single method even deep learning one \cite{Hajirahimi}. This combination is not addressed in our paper, however, the experiments of this paper reveal that the single models are the best choice for a user who would like to promptly estimate an energy task. In contrast, hybrid models are the best choice for users who know the basics of machine learning techniques and can integrate more than a model with a suitable optimization algorithms. Nevertheless, keep in mind that, the growing number of parameters that yield after combining one or more models is the most negative aspects of such hybridization. 
\end{enumerate}

\begin{table*}[ht]
\caption{Analytical comparison among the selected models} \label{Overall_Comparison}
\centering
\begin{tabular}{llll}
\hline
No. & Method & Advantages & Disadvantages \\ \hline
   1& ARIMA     & \begin{minipage}[t]{0.4\textwidth}
\begin{enumerate}

\item Good performance for short-term data and UTS 
\item Ability to handle non-stationary TS data
\item Require fewer training data samples
\end{enumerate}
\end{minipage}& \begin{minipage}[t]{0.4\textwidth}
\begin{enumerate}

\item Linear and parametric method
\item  Poor performance for long-term and MTS problems
\item Suffer with trend and seasonality
\end{enumerate}
\end{minipage}\\ 
\hline

   2 & SARIMA     &  \begin{minipage}[t]{0.4\textwidth}
\begin{enumerate}
\item  Ability to handle TS data with seasonality
\end{enumerate}
\end{minipage}
& \begin{minipage}[t]{0.4\textwidth}
\begin{enumerate}
\item  Supporting seasonality adds more parameters
\end{enumerate}
\end{minipage} \\ \hline

   3 &  VAR    &  \begin{minipage}[t]{0.4\textwidth}
\begin{enumerate}
\item  Good performance when solve MTS problems
\item Ability to measure the explanatory change on the variables over time
\item  Support autocorrelation
\item Require fewer training data samples
\end{enumerate}
\end{minipage}   &  \begin{minipage}[t]{0.4\textwidth}
\begin{enumerate}
\item   Linear and parametric method 
\item  High computation cost, where many coefficients must be estimated. 
\item Possibility of false correlations (BVAR is a solution)
\item Poor performance with long lag observations
\end{enumerate}
\end{minipage}\\ \hline

   4 &  ES    &  \begin{minipage}[t]{0.4\textwidth}
\begin{enumerate}
\item  Simple and strong mathematical foundation
\item Require fewer training data samples
\item Based on exponential decay of past observations weights
\end{enumerate}
\end{minipage}   & \begin{minipage}[t]{0.4\textwidth}
\begin{enumerate}
\item  Linear and parametric method 
\item Suffer with trend, seasonality, and long-term TS data
\item No support for autocorrelation
\end{enumerate}
\end{minipage} \\ \hline
   5 &  MES    & \begin{minipage}[t]{0.4\textwidth}
\begin{enumerate}
\item  Ability to solve MTS problems
\end{enumerate}
\end{minipage}    & \begin{minipage}[t]{0.4\textwidth}
\begin{enumerate}
\item  Longer training time than original ES
\end{enumerate}
\end{minipage}\\ \hline

   6 &    SES  &   \begin{minipage}[t]{0.4\textwidth}
\begin{enumerate}
\item   Ability to process data with trend and seasonality
\item Simple computation than original ES
\end{enumerate}
\end{minipage}  & \begin{minipage}[t]{0.4\textwidth}
\begin{enumerate}
\item   Supporting seasonality adds a smoothing parameter
\end{enumerate}
\end{minipage}\\ \hline

   7 & kNN     &   \begin{minipage}[t]{0.4\textwidth}
\begin{enumerate}
\item   Simple and efficient for small datasets
\item Efficient for trend and seasonality in TS data 
\item Data distribution should not be known in advance
\end{enumerate}
\end{minipage}  & \begin{minipage}[t]{0.4\textwidth}
\begin{enumerate}
\item Poor performance with large datasets,
\item High computation cost for large datasets
\item Optimum distance metric is unknown
\end{enumerate}
\end{minipage} \\ \hline
   8&   SVR   &  \begin{minipage}[t]{0.4\textwidth}
\begin{enumerate}
\item Good performance with long-term TS 
\item No overfitting and can avoid trapping in a local minima
\item Introduce nonlinearity using a kernel function
\end{enumerate}
\end{minipage}   &  \begin{minipage}[t]{0.4\textwidth}
\begin{enumerate}
\item Poor performance with short-term large-scale dataset
\item Select some sparse support vectors to model
\item Slowness due to calculating the  epsilon-insensitive loss function
\end{enumerate}
\end{minipage} \\ \hline
\end{tabular}

\end{table*}

\begin{table*}[ht]
\caption*{} \label{Overall_Comparison}
\centering
\begin{tabular}{llll}
\hline
No. & Method & Advantages & Disadvantages \\ \hline

   9 &   LSSVR   &   \begin{minipage}[t]{0.4\textwidth}
\begin{enumerate}
\item Faster than the original SVR
\end{enumerate}
\end{minipage}  & \begin{minipage}[t]{0.4\textwidth}
\begin{enumerate}
\item Poor performance with chaotic properties of TS data.
\end{enumerate}
\end{minipage}\\ \hline

   10 &   MLP   &  \begin{minipage}[t]{0.4\textwidth}
\begin{enumerate}
\item Simple training requirements
\item perform multi-step ahead forecasts
\item Ability to handle multivariate inputs. 
\item Efficient learning from TS lag observations 
\end{enumerate}
\end{minipage}   & \begin{minipage}[t]{0.4\textwidth}
\begin{enumerate}
\item  Linearity and slow convergence
\item High possibility to trapped into local minima
\item Sensitivity to learning rate
\item Poor performance in case of stochastic data. 
\end{enumerate}
\end{minipage}\\ \hline
   11 &   RNN   &    \begin{minipage}[t]{0.4\textwidth}
\begin{enumerate}
\item  Work Good even in case of noisy and missing data
\item Robust and stable in all forecasting horizons
\item Parameter sharing over the network and store information for further use
\end{enumerate}
\end{minipage}  &    \begin{minipage}[t]{0.4\textwidth}
\begin{enumerate}
\item Vanishing the gradient in long-term data
\item Unsuitable for long-term forecasting
\item High computational cost due to intensive connections among cells
\end{enumerate}
\end{minipage}\\ \hline
   12 &  LSTM    &  \begin{minipage}[t]{0.4\textwidth}
\begin{enumerate}
\item  Suitable for long-term forecasting 
\item Solve the gradient vanishing problem of RNN
\end{enumerate}
\end{minipage}   & \begin{minipage}[t]{0.4\textwidth}
\begin{enumerate}
\item Sensitive to random initialization for weight
\item High computation cost (Big no. of memory cells linked to the size of the recurrent weight matrices)
\end{enumerate}
\end{minipage}\\ \hline
   13 & ESN    &   \begin{minipage}[t]{0.4\textwidth}
\begin{enumerate}
\item  Simple training procedure
\item Efficient and low cost computation
\end{enumerate}
\end{minipage}   & \begin{minipage}[t]{0.4\textwidth}
\begin{enumerate}
\item Show similar performance of RNN
\end{enumerate}
\end{minipage}\\ \hline
   14 & DRNN     &   \begin{minipage}[t]{0.4\textwidth}
\begin{enumerate}
\item  Powerful self-learning and generalization abilities
\item Ability to handle nonlinear and heterogeneous data
\item Good performance with large datasets and complex initial features
\end{enumerate}
\end{minipage}  &  \begin{minipage}[t]{0.4\textwidth}
\begin{enumerate}
\item  High computational cost and slow convergence.
\item Cumbersome process of tuning parameter 
\item Due to deep structure, classical learning algorithms suffer to find optimal solution
\end{enumerate}
\end{minipage}  \\ \hline
   15 & DLSTM & \begin{minipage}[t]{0.4\textwidth}
\begin{enumerate}
\item The integrated memory loops make fast learning
\item Support layer-wise based learning model
\end{enumerate}
\end{minipage} & \begin{minipage}[t]{0.4\textwidth}
\begin{enumerate}
\item High computational cost and slow convergence 
\item Need plentiful training data
\end{enumerate}
\end{minipage} \\\hline

\end{tabular}

\end{table*}
\end{document}